\documentclass[dvipsnames,format=sigconf,anonymous=false,review=false,nonacm]{acmart}
\usepackage{dcolumn}

\newcommand{\rc}[1]{{#1}}

\usepackage{algorithm,algpseudocode}
\usepackage{stfloats}
\usepackage{adjustbox}
\usepackage{subfigure}
\usepackage{multirow,enumitem}

\AtBeginDocument{%
  \providecommand\BibTeX{{%
    \normalfont B\kern-0.5em{\scshape i\kern-0.25em b}\kern-0.8em\TeX}}}

\begin{document}

\title{Learning from Offline and Online Experiences:\\ A Hybrid Adaptive Operator Selection Framework}

\author{Jiyuan Pei}
\email{peijy2020@mail.sustech.edu.cn}
\orcid{0000-0001-9860-5160}
\affiliation{%
  \institution{Southern University of Science and Technology}
  \city{Shenzhen}
  \country{China}
}
\affiliation{%
  \institution{Victoria University of Wellington}
  \city{Wellington}
  \country{New Zealand}}

\author{Jialin Liu}
\authornote{Corresponding author}
\email{liujl@sustech.edu.cn}
\orcid{0000-0001-7047-8454}
\affiliation{%
  \institution{Southern University of Science and Technology}
  \city{Shenzhen}
  \country{China}
}

\author{Yi Mei}
\email{yi.mei@ecs.vuw.ac.nz}
\orcid{0000-0003-0682-1363}
\affiliation{%
  \institution{Victoria University of Wellington}
  \city{Wellington}
  \country{New Zealand}}

\begin{abstract}
    In many practical applications, usually, similar optimisation problems or scenarios repeatedly appear. Learning from previous problem-solving experiences can help adjust algorithm components of meta-heuristics, e.g., adaptively selecting promising search operators, to achieve better optimisation performance. However, those experiences obtained from previously solved problems, namely \emph{offline experience}s, may sometimes provide misleading perceptions when solving a new problem, if the characteristics of previous problems and the new one are relatively different.
    Learning from \emph{online experience}s obtained during the ongoing problem-solving process is more instructive but highly restricted by limited computational resources.
    This paper focuses on the effective combination of offline and online experiences. A novel hybrid framework that learns to dynamically and adaptively select promising search operators is proposed.
    Two adaptive operator selection modules with complementary paradigms cooperate in the framework to learn from offline and online experiences and make decisions. An adaptive decision policy is maintained to balance the use of those two modules in an online manner.
    Extensive experiments on 170 widely studied real-value benchmark optimisation problems and a benchmark set with 34 instances for combinatorial optimisation show that the proposed hybrid framework outperforms the state-of-the-art methods.
    Ablation study verifies the effectiveness of each component of the framework.
\end{abstract}


\maketitle

\section{Introduction}

Meta-heuristics are widely used for solving various optimisation problems in real-world applications \cite{metaheuristic_survey2018,beheshti2013review}, as they exhibit promising capabilities in finding sub-optimal solutions of complex optimisation problems within a relatively limited computational budget. 
Traditionally, the development of new meta-heuristics highly relies on manual efforts to address specific, unstudied characteristics of new problems in both research and practical applications. Real-world optimisation problems typically have unpredictable characteristics, requiring additional labour for configuring algorithm components when unexpected characteristics arise. In response to these challenges, there is a growing trend in leveraging the impressive capabilities of machine learning methods to assist algorithm configuration.  Some learning-assisted optimisation methods aim to automatically adjust the configuration of meta-heuristics, such as dynamically selecting search operators, to better align with the requirements of different optimisation stages, by learning from previous problem-solving experiences~\cite{Schede_2022,aos_thesis_2010:fialho}. An intuitive example is that, by learning the relation between the performance of available search operators and features of different optimisation states, the most suitable operator for each state can be selected and utilised~\cite{GROLIMUND1997326}. Methods that improve the optimisation ability of meta-heuristics by learning-assisted operator selection are widely studied in the research community of Adaptive Operator Selection (AOS) \cite{9712324,Lu2020A,Consoli_2016} and selection hyper-heuristics \cite{DRAKE2020405,QIN2021107252,16051646720221201,10.1145/3449726.3463187}.

Commonly in practice, a series of problems under a specific formulation are required to be solved sequentially.
Rich experiences can be gained from optimisation processes of previous problems and learned to better solve the new problem. Such experience can be referred to as \emph{offline experience} as it is collected before handling the new problem.
For example, a delivery service provider needs to make delivery plans every day for the same city as the daily orders are different and hard to predict. 
Past problem-solving processes, plans and performance build the offline experience, which are collected and learned without consuming the limited computational resources for making today's plan.
However, one main challenge associated with learning from offline experience is that the learned knowledge may sometimes be misleading when the new problems to be solved are relatively different from past ones.
In such scenarios, the optimisation performance is not boosted but corrupted by the learned model. 
In other fields than AOS, some works measure the similarity between past problems and new ones during the optimisation process~\cite{contreras2014blind} or even predicting characteristics of new problems~\cite{Mandi_s_2020}, however, the high computational cost associated with the accurate measurement and the difficulty in accurate prediction are not negligible.
Exploiting the effective use of offline experience for solving new problems while mitigating the resultant, additional cost in the optimisation process is essential.

On the other hand, new experience can be gained while solving the new problem, namely \emph{online experience}. Meta-heuristics optimise by iteratively applying search operators on current solutions to generate new solutions. Naturally, the online experience collected during past iterations can be employed to guide the subsequent iterations. 
Notwithstanding the available online experience, the limited computational resources become one major obstacle in its best use. 
Increasing resources for online learning leads to a decrease in resources for searching. 
As many machine learning methods, especially deep learning ones, are computationally expensive, resources allocated for search and online learning should be carefully balanced. 
Furthermore, data collected during search is also significantly less than that can be collected from solving processes of past problems.  
This strongly restricts the utilisation of complex learning models that require a great amount of data.

Specific to AOS, various works attempt online experience learning~\cite{10.1007/978-3-642-15844-5_20,10.1145/1068009.1068251,Consoli_2016,JIAO2023101225} and offline experience learning~\cite{10.1007/978-3-319-50349-3_13,10.1145/3321707.3321813,Lu2020A}.
It is intuitive to combine offline and online experiences.
However, the combination is currently underdeveloped. To the best of our knowledge, within the literature of AOS, the studies of \cite{a15010024} and \cite{aydin2023adaptive} are the only ones that learn from historical and current problem-solving processes. 
The work of~\cite{a15010024} introduces a reinforcement learning-based AOS method and conducts offline learning on the problem instance identical to the one used for online learning and testing.
The subsequent study~\cite{aydin2023adaptive} further investigates the effectiveness of learning offline experiences of solving different problem instances and types. Experiments indicate that the offline experiences are promising for solving identical problem instances~\cite{aydin2023adaptive}. However, when the characteristics of problems or instances are relatively different, the performance of the learned model is limited. Offline experiences may be seriously misleading for solving new problems. In this situation, simply updating the RL policy model online is not sufficient. 
Additional techniques for adjusting the model or decisions online are needed for selecting promising operators.

To address the above issues, this paper proposes a hybrid AOS framework that combines two complementary modules to effectively learn offline and online experiences. Specifically, the contribution of this paper is summarised as follows.
\begin{itemize}
\item A hybrid AOS framework with the following components that learns from offline and online experiences is proposed.
\begin{enumerate}[label=(\roman*)]
    \item A state-based AOS module learns the mapping from the optimisation state to promising operator selection. It firstly learns offline experiences from past solved problems, then continuously learns online experiences.
    \item A stateless AOS module learns only from online experiences with a relatively low computational cost.
    \item A decision policy balances the aforementioned two modules and selects an operator.
\end{enumerate}
\item The effectiveness and generalisation of our proposed hybrid framework are validated by extensive experiment studies on 170 commonly studied real-value optimisation problems and 34 instances of a challenging constrained combinatorial optimisation problem known as the Capacitated Vehicle Routing Problem with Time Windows (CVRPTW)~\cite{10.1287/opre.35.2.254}, compared to the state-of-the-art AOS methods and their variants equipped with our online experience learning module. 
Ablation study shows the unique contribution of each component in our hybrid framework.
\end{itemize}

The rest of the paper is organised as follows. Section \ref{sec:background} introduces related work. Section \ref{sec:proposed_method} describes the hybrid AOS framework proposed in this paper. Experimental studies are presented and discussed in Section \ref{sec:experiment}. Section \ref{sec:conclusion} concludes.

\section{Background\label{sec:background}}

This section presents the research status of adaptive operator selection, and studies related to learning from experiences to assist optimisation on different problems.

\subsection{Adaptive Operator Selection\label{sec:AOS}}

AOS focuses on dynamically and adaptively selecting the most suitable operator during the optimisation process of meta-heuristics~\cite{10.5555/93126.93146,aos_thesis_2010:fialho}. 
Based on whether the current search state is considered in decision-making or not, existing AOS methods in literature can be categorised into two paradigms: stateless methods and state-based methods~\cite{10.1145/2598394.2598451,10.1007/978-3-031-20862-1_41}. 
Stateless AOS methods make decisions based on operators' historical performance during the current optimisation process, e.g., they tend to select the one with the best historical performance. Although a stateless AOS method can learn from offline and/or online experiences, to the best of our knowledge, existing stateless AOS methods utilise online experiences only \cite{aos_thesis_2010:fialho,10.1145/1068009.1068251,6410018,KALATZANTONAKIS2023118812}.
State-based AOS methods learn the mapping between search state features and the best operators in the state. \rc{State features provide information related to operators' performance, including features of current solutions (e.g., objective values, diversity measurements, etc.), statistics about optimisation process (e.g., budget left, number of iterations without improvement, etc.), and information about the optimisation problem~\cite{9852781,10.1145/3321707.3321813,Kletzander_Musliu_2023}.} State-based AOS methods can learn both offline and online experiences from different runs and different problems, in which the state spaces need to be identical.

In the study of stateless AOS, the selection of operators is usually formulated as the multi-armed bandit problem in dynamic scenarios~\cite{10.1145/1068009.1068251,aos_thesis_2010:fialho,LAGOS202470,8477930,KALATZANTONAKIS2023118812}. Each operator owns a relatively stable ability to generate good solutions in a given period during optimisation.
The ability of an operator is estimated by the historical credits it gained. The operator with the best estimated ability will be assigned the largest probability of being selected, while others are still explored with a relatively low probability.
Commonly, a stateless AOS method is composed of two components \cite{aos_thesis_2010:fialho}, Credit Assignment (CA) which assesses the credit, and Operator Selection Rule (OSR) which calculates the selection probability of each operator.

Distinguished from stateless AOS methods, state-based AOS methods incorporate the current search state into the decision-making process. Training the mapping from search state to operator selection is an essential part of state-based AOS methods. It usually follows the reinforcement learning paradigm \cite{9852781,10.1145/3321707.3321813}. The iterative operator selection during optimisation is defined as a Markov decision process~\cite{10.1145/2598394.2598451}. Each operator is an action to be selected at each state during the search.
The search state transforms when new solutions are generated by operators.
The reward function assesses each operator application. Elaborated high-dimensional state features are proposed for a comprehensive description of states during the search process, leading to the tendency to use deep neural networks as the policy model~\cite{10.1145/3321707.3321813,9852781}. Popular deep reinforcement learning (DRL) methods, such as Double Deep Q-Network (DDQN) \cite{Hasselt_Guez_Silver_2016} and Proximal Policy Optimization (PPO) \cite{SchulmanWDRK17}, are employed in the training of state-based AOS.
The effectiveness of DRL methods in state-based AOS has been validated across a diverse spectrum of optimisation problems and meta-heuristics \cite{10.1145/3321707.3321813,Lu2020A,9712324,9852781}.

Problems that own similar characteristics to the given test problems are usually selected for training in studies of offline-trained AOS.
For example, in the study of \cite{10.1145/3321707.3321813} tested on the CEC2005 benchmark~\cite{suganthan2005problem}, the training set and test set both contain functions of the same classes.
In the study of \cite{Lu2020A}, the training and test problems are generated with the same distribution. 
However, in real-world applications, the new problem is often unseen. Thus, the assumption of following the same distribution or problem similarity can hardly be guaranteed. 
If the training and test problems own a relatively large difference in characteristics, the learned offline experiences may be seriously misleading when solving the test problems.

In real-world applications, commonly, new problems at hand have relatively different characteristics from the problems that were previously solved. However, this scenario is not considered in most studies of state-based AOS.
Continuously updating the offline-learned policy with online experiences of solving the test problem is a potential way to address the issue. The studies of \cite{a15010024} and \cite{aydin2023adaptive} are the only two known to investigate learning with offline experiences from training problems and online experiences from the test problem. However, an identical problem instance was used as training and test data in the work of \cite{a15010024}. Their recent study \cite{aydin2023adaptive} further investigates transferring the learned model to the online learning AOS method between different problem instances and types. Experiments suggest that the transferability is promising for solving identical problem instances, thus the testing instance is also the training instance \cite{aydin2023adaptive}. However, when the problem characteristics are relatively different, the performance of transferring the learned model is limited \cite{aydin2023adaptive}. Offline experiences may be seriously misleading in the subsequent problem-solving process. In such situation, simply updating the RL policy model online is not sufficient. It requires further method development to handle the potential misleading and assist the operator selection.

Stateless AOS process typically involves only 
simple scalar operation, resulting in a considerably low computational cost. State-based AOS, particularly DRL-based methods, involves search states in decision-making and exhibits superior performance in complex problems.  
The training of policy, e.g., neural networks in DRL-based methods, consumes substantial computation resources. When solving new problems, the online training of state-based AOS model is not sufficient at the early stages, stateless AOS methods can provide a relatively sound decision at a low cost. In this respect, stateless and state-based AOS methods emerge as complementary strategies.

\subsection{Generalisation in Optimisation with Learning}
The performance of a trained machine learning model on unseen data depends on its generalisation ability. 
In the context of learning to solve optimisation problems,
learning model's generalisation ability affects the optimisation performance on unseen problems.
Though the generalisation ability of trained models is commonly tested in studies of offline-trained AOS, there remains a gap in specific approaches focusing on improving the generalisation ability.

Learn-to-optimise (L2O) is a related topic of AOS
\cite{Tang2024L2Oreview,JMLR:v23:21-0308,shen2021learning}.
Not only to configure meta-heuristics, many L2O methods directly train the learning model to output high-quality solutions for a given problem. Training of models with high generalisation performance gains increasing attention in L2O studies. The work of \cite{NEURIPS2022_ca70528f} investigates improving the model's generalisation ability with knowledge distillation approaches. A student model is trained with the assistance of multiple teacher models. Each teacher model learns from problems generated by a unique distribution, so that the student is able to handle multiple problem instance distributions. In the work of \cite{zhou2023omnigeneralizable}, an omni-generalisable method is proposed. A meta-model is firstly trained on a given set of problem instance distributions. Then, to solve a specific instance distribution, a relatively small number of instances are sampled with the distribution and used to fine-tune the model. The study of \cite{NEURIPS2020_51f4efbf} proposes a L2O method with curriculum Learning to adaptively select training problems. The model trained with the curriculum learning method demonstrates better generalisation ability on a range of problems. In summary, the existing studies are still based on knowledge of problem instances' characteristics, and require a set of problem instances that own similar characteristics to the test instance for training. In operator selection tasks with the scenario that characteristics of test problems are relatively different from training problems but unknown, the applicability of these methods is highly limited.

Research of transfer learning for optimisation study transfers the knowledge obtained from one problem domain to another \cite{8114198,9134370}, is another related topic of this paper. The efficiency and effectiveness of learning on the target problem domain benefit from the knowledge, e.g., collected data or trained model, obtained from the source problem domain. However, transfer learning studies often take the transferred knowledge from the source domain to help retrain the model on the target domain offline, with one or multiple pre-known instances in the target problem domain. It is also not applicable in the scenario focused on in this paper, as
pre-known instances in the target problem domain are unavailable.

\section{Hybrid Adaptive Operator Selection Framework\label{sec:proposed_method}}

\begin{figure*}[htbp]
    \centering
    \includegraphics[width=.85\linewidth]{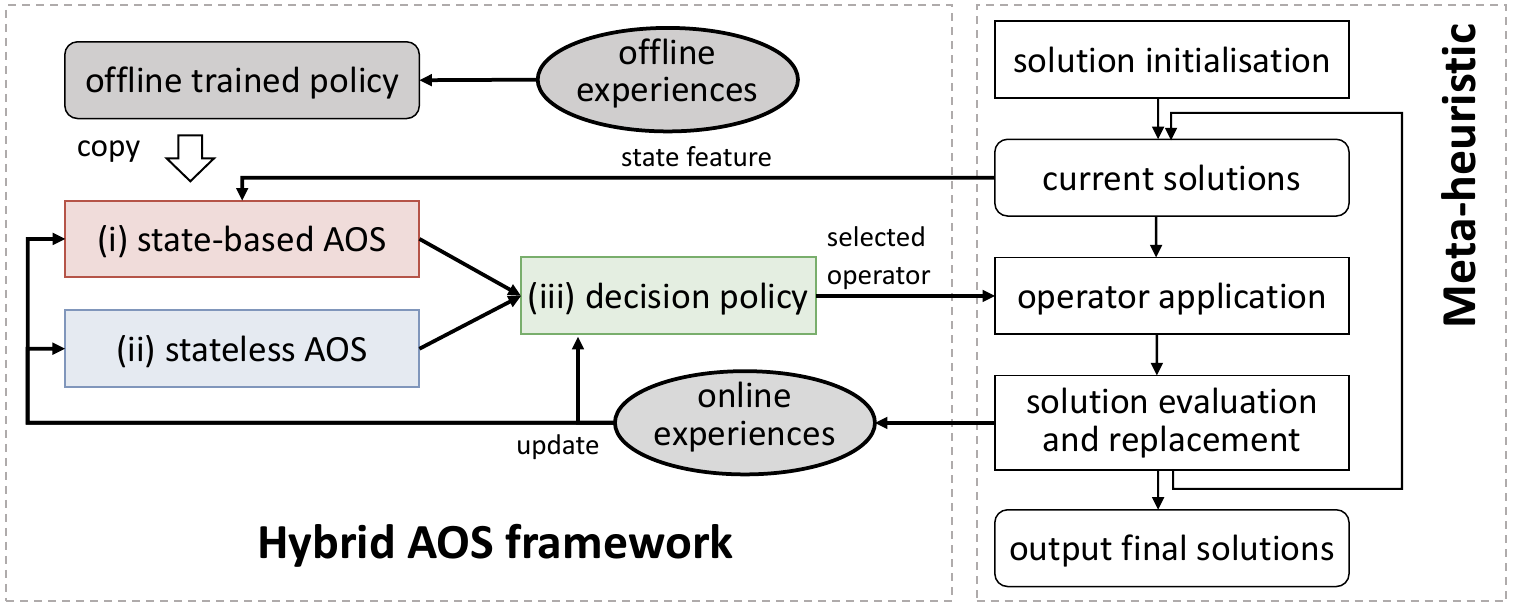}
    \caption{The proposed hybrid AOS framework.}
    \label{fig:framework}
\end{figure*}

In this paper, we propose a hybrid AOS framework designed to maximise the utilisation of offline and online experiences to boost the optimisation of new problems. 
Fig. \ref{fig:framework} illustrates the main components of our framework and how it interacts with a meta-heuristic.
The three main components include (i) a state-based AOS module in which the policy model is trained on offline experiences of optimising other problems (instances) before and online experiences collected during the ongoing optimisation of the current problem (instance); (ii) a stateless AOS module that learns from online experiences with low computational cost; and (iii) a decision policy adjusting mechanism that adjusts the weights of using the two AOS modules in decision-making.
The state-based AOS module plays a pivotal role in learning experiences, while the stateless AOS module and the decision policy adjusting are embedded in the framework to address the potential misleading from offline experiences.

Algorithm \ref{algo:framework} details how a meta-heuristic cooperates with the proposed hybrid AOS framework for optimising a minimisation problem. Notably, Algorithm \ref{algo:framework} serves as a general demonstration. Details in each component, e.g., the required inputs or training method, may vary according to the specific methods used.
First, the state-based AOS model is trained offline with experience (i.e., data) collected in past optimisation of other problems or instances, such as the state features, the operator selected and the objective value improvement achieved by the selected operator in each iteration. Then, the offline trained model $\pi$ is taken as the initial model for online learning and is updated iteratively when the hybrid framework is conducted to solve the problem at hand. 
At the starting stage of the current optimisation, each component is initialised (lines \ref{line:init1}-\ref{line:init2}). At each optimisation iteration, one of the state-based and stateless modules is applied with probability $p$ to select an operator (lines \ref{line:decision1}-\ref{line:decision2}), which is then applied to the current solution for generating new ones (line \ref{line:ope}). 
Then, the new solution is evaluated (line \ref{line:ef}).
The stateless AOS module is updated by credit assignment and enriching credit $record$ (lines \ref{line:CA}, \ref{line:SLupdate}) using the objective values of the original and the new solution.
The features of the new optimisation state are extracted (line \ref{line:state}) and used to update the state-based AOS model (line \ref{line:SBupdate}). Then, the probability $p$ is adaptively adjusted based on whether an improvement is achieved by the application of the selected operator or not (lines \ref{line:wupdate1}-\ref{line:wupdate2}). 

\begin{algorithm}[htbp]
\caption{Meta-heuristic with the hybrid framework\label{algo:framework}}
\begin{algorithmic}[1]
\Require{a set of $K$ operators $\mathcal{A} = \{a_1,\dots,a_K\}$, a problem to be solved and the corresponding solution evaluator $f()$, maximal iteration number $T$, offline trained state-based model $\pi$, and optimisation state feature extraction method $SE()$, reward function in state-based training $Reward()$, a stateless AOS method consisting of $CA()$ and $OSR()$, upper bound and lower bound of probability $p_u \in (0,1)$ and $p_l \in (0,p_u)$ }
\State $p \leftarrow p_u$ \label{line:init1}
\State $x_0 \leftarrow$ random solution initialisation 
\State $y_0 \leftarrow f(x_0)$ \Comment{\emph{solution evaluation}}
\State $s_{0} \leftarrow SE(x_{0},f_0)$\label{line:state_init} \Comment{\emph{state feature extraction}}
\State $record \leftarrow$ $\{\}$ \label{line:init2}
\For{$t \leftarrow 1$ to  $T$} \label{line:loop1} 
    \If{$random(0,1) < p$} \label{line:decision1}
         \State $a^*_{t} \leftarrow OSR(record,\mathcal{A})$\label{line:SL} \Comment{\emph{stateless selection}}
    \Else
        \State $a^*_{t} \leftarrow \pi(s_{t-1},\mathcal{A})$  \Comment{\emph{state-based selection}}\label{line:SB}
    \EndIf \label{line:decision2}
    \State $x_{t} \leftarrow$ apply operator $a^*_{t}$ to $x_{t-1}$ \label{line:ope}
    \State $y_{t} \leftarrow f(x_{t})$ \label{line:ef}
    \State $c_{t} \leftarrow CA(y_{t}, y_{t-1})$ \label{line:CA} \Comment{\emph{credit assignment}}
    \State add $ (a^*_{t},c_{t})$ into $record$ \Comment{\emph{update record}}\label{line:SLupdate}
    \State $s_{t} \leftarrow SE(x_{t},y_{t})$ \Comment{\emph{feature extraction}}\label{line:state} 
    \State $r_{t} \leftarrow Reward(y_{t}, y_{t-1})$ \Comment{\emph{calculate reward}}\label{line:reward}
    \State update $\pi$ with $(s_{t-1},s_{t},a_{t},r_{t})$\label{line:SBupdate} \Comment{\emph{update model}}
    \If {$y_{t} < y_{t-1}$}\label{line:wupdate1}
        \State $p \leftarrow \frac{p+p_u}{2}$ \label{line:w+}
    \Else
        \State $p \leftarrow \frac{p+p_l}{2}$\label{line:w-}
    \EndIf \label{line:wupdate2}
\EndFor\label{line:loop2}
\State \Return $x_T$
\end{algorithmic}
\end{algorithm}

The state-based AOS module utilises the trained model to make decisions. The model takes features of the current search state as the input, and outputs either the estimated performance of each operator or the estimated optimal operator selection probability distribution. In the former case, the operator with the maximal estimated performance is commonly deterministically selected. In the latter case, operators are randomly selected based on the probability distribution. 
After applying the selected operator, a reward value is calculated based on the pre-defined reward function (line \ref{line:reward}). Subsequently, the policy model is updated with the selected operator, the reward value, the state features before and after the operator application (line \ref{line:SBupdate}). 
On the other hand, the stateless AOS module makes decisions based on the history of the optimisation process instead of the current search state features. The historical performance of operators in each iteration is recorded (line \ref{line:SLupdate}), and the one with the largest historical credits is assigned with the most selection probability. The CA method calculates the credit gain by applying the selected operator (line \ref{line:CA}). 
The process of the stateless AOS module involves only simple scalar addition and multiplication operations, avoiding the need for expensive model training. As a result, the additional computation cost introduced by a stateless AOS module is negligible.

The state-based and stateless AOS modules make decisions independently,  
while experiences obtained subsequently to decisions of one module are utilised to update both modules. 
In this case, their complementary decision strategies are maintained respectively and all valuable online experiences are fully utilised. Each operator selection in solving the current problem is not only a sampling for online learning but also a step of optimisation. An appropriate way to make the overall decision is essential to push the optimisation process effectively and obtain instructive experiences for online learning. The adaptive adjusted decision policy (lines \ref{line:decision1}-\ref{line:decision2}, \ref{line:wupdate1}-\ref{line:wupdate2}) in the proposed framework works to provide the overall decision.

The proposed decision policy is developed based on the relation of stateless AOS methods and local optima met during optimisation \cite{10.1007/978-3-031-20862-1_41,10.1145/3583131.3590399}.
The study of \cite{10.1007/978-3-031-20862-1_41} investigates the behaviour of  AOS methods on a complex combinatorial optimisation and suggests that the performance of stateless AOS methods may be promising when the current search area of solution space is away from the local optima, but may be relatively poor when local optima are close. An effective method for improving AOS performance by predicting local optima \cite{10.1145/3583131.3590399} verifies the relation. According to the observed relation, the weight adjusting mechanism is designed to reduce the weight of the stateless module when no improvement is achieved (lines \ref{line:w+}), i.e., a local optimum is likely to be close, and increase otherwise (lines \ref{line:w-}). Commonly during the optimisation process of a meta-heuristic, the frequency of meeting local optima gradually increases as the quality of solutions improves. At the same time, the performance of the online trained state-based AOS model also increases as the training data gradually increases. The proposed weight adjusting mechanism increases the weight assigned to the stateless method when it performs well, and reduces the weight for the stateless method when its performance decreases and the performance of the state-based AOS model increases, achieving an appropriate balance between the two.

The focus of this paper is to develop the AOS hybrid framework that combines complementary component AOS methods to effectively learn both offline and online experiences, rather than proposing specific AOS methods, either in the stateless or the state-based paradigm. 
The hybrid framework is general.
Existing stateless AOS methods (e.g., methods based on multi-armed bandit algorithms \cite{aos_thesis_2010:fialho,LAGOS202470,8477930,KALATZANTONAKIS2023118812}) and state-based AOS methods (e.g.,  reinforcement learning-based AOS methods \cite{a15010024,aydin2023adaptive} and supervised learning-based AOS methods \cite{Consoli_2016,Geng_2022}) can be applied as the two modules.
To apply an existing state-based AOS method as the state-based module, the hybrid framework does not require modifying the existing offline training process. State features for decision-making should be identical in both offline and online experiences.

\section{Experimental study\label{sec:experiment}}
Our proposed hybrid framework is applied to two main optimisation domains, real-value optimisation and combinatorial optimisation. 
For real-value optimisation, the CEC2005 \cite{suganthan2005problem}, CEC2017 \cite{wu2017problem} and BBOB2023 \cite{nikolaus_hansen_2019_2594848}
single objective benchmark functions with different dimensions are selected. For combinatorial optimisation, CVRPTW with widely used Solomon benchmark instances \cite{10.1287/opre.35.2.254} are used.

Three sets of experiments are conducted to comprehensively evaluate the hybrid AOS framework (denoted as HF):
\begin{enumerate}
    \item To verify the effectiveness of HF, we compare it with three state-of-the-art methods: state-based AOS methods on real-value problems and CVRPTW respectively, DE-DDQN~\cite{10.1145/3321707.3321813} for real-value optimisation, DQN-GSF~\cite{9852781} for combinatorial optimisation, and a stateless AOS method (denoted as SL) composing a classic OSR (adaptive pursuit~\cite{10.1145/1068009.1068251}) and a classic CA (fitness improvement rate~\cite{6410018}). Uniformly random selection (denoted as Random) is also compared.
    \item To verify the effectiveness of the online experience learning of the two modules in HF, we conduct ablation studies by comparing: HF, HF without online updating the state-based AOS model (denoted as HF-NU), and two state-of-the-art state-based AOS methods with online updating (denoted as DE-DDQN-U and DQN-GSF-U).
    \item To verify the effectiveness of the decision adjusting, HF is compared with different fixed module selection probability $p=0.5$, 0.3 and 0.1, denoted as HF-NA(0.5), HF-NA(0.3) and HF-NA(0.1), respectively.
\end{enumerate}
Table \ref{tab:method_notation} details the notations of AOS methods compared in experiments and highlights their differences.

\subsection{Experiment Setting\label{sec:experiment_setting}}

\begin{table*}[htbp]
    \centering
    \caption{Compared AOS methods and their components.}
    \label{tab:method_notation}
    \setlength{\tabcolsep}{2pt}
    \begin{tabular}{cl|cc|c|c|p{0.36\linewidth}}
\toprule
\multicolumn{2}{c|}{\multirow{2}{*}{Method}}  & \multicolumn{2}{c|}{State-based Module} & Stateless & Decision Policy &  \multicolumn{1}{c}{\multirow{2}{*}{Description}} \\
&     & Offline Training      & Online Updating       &   Module    &   Adjusting  &      \\
\midrule
\textbf{Ours} & \multicolumn{1}{c|}{\textbf{HF}}         & $\checkmark$ & $\checkmark$ & $\checkmark$ & $\checkmark$ & Our proposed framework  \\
\midrule
\multirow{2}{*}{Ablation} & \multirow{2}{*}{HF-NU}      & \multirow{2}{*}{$\checkmark$} &       & \multirow{2}{*}{$\checkmark$} & \multirow{2}{*}{$\checkmark$} & Our proposed framework without online updating the state-based AOS model \\
\cline{2-7}
\multirow{2}{*}{Study}  & \multirow{2}{*}{HF-NA($p$)} & \multirow{2}{*}{$\checkmark$} & \multirow{2}{*}{$\checkmark$} & \multirow{2}{*}{$\checkmark$} &       & Our proposed framework without decision policy adjusting. Decision weight is fixed as $p$\\
\midrule
 & \multirow{2}{*}{DE-DDQN}        & \multirow{2}{*}{$\checkmark$} &       &       &       & State-based AOS method for solving real-value problems~\cite{10.1145/3321707.3321813}\\
\cline{2-7}
State-of- & DE-DDQN-U       & $\checkmark$ & $\checkmark$ &       &       & Variant of DE-DDQN~\cite{10.1145/3321707.3321813} with online updating\\
 \cline{2-7}
the-arts & DQN-GSF        & $\checkmark$ &       &       &       & State-based AOS method for solving CVRPTW~\cite{9852781}\\
\cline{2-7}
 & DQN-GSF-U       & $\checkmark$ & $\checkmark$ &       &       & Variant of DQN-GSF~\cite{9852781} with online updating\\
\cline{2-7}
& SL         &       &       & $\checkmark$ &       & Stateless AOS method   \cite{6410018,10.1145/1068009.1068251} \\
\midrule
& Random     &       &       &       &       & Uniformly random selection\\
\bottomrule
    \end{tabular}
\end{table*}

\paragraph{Training and test sets}
For real-value problems, this paper follows the setting of \cite{10.1145/3321707.3321813} and takes the 16 functions with dimensions 10 and 30 in the CEC2005 benchmark set \cite{suganthan2005problem} for training, composing 32 training problems.
To simulate the scenario that characteristics of unseen future problems are relatively different from training problems, 30 functions of the CEC2017 benchmark \cite{wu2017problem} with dimensions 10, 30 and 50, together with 24 functions of the BBOB2023 \cite{nikolaus_hansen_2019_2594848} with dimensions 10 and 20 are used for testing, composing in a total of 138 test problems. 
In the Solomon CVRPTW instance set \cite{10.1287/opre.35.2.254}, there are three types of instances generated with three different distributions, including C-type (C101-C109, C201-C208), R-type (R101-R112, R201-R211) and RC-type (RC101-RC108, RC201-RC208). R101 is used as the training instance, following the generalisation experiment setting in \cite{9852781}. C-type instances and RC-type instances are used as test instances, with a total number of 33. 

\paragraph{Evaluation metric}
All tested problems are minimisation problems, where solutions with smaller objective values is better.
The involved meta-heuristics are population-based stochastic algorithms. To perform a fair comparison, 30 populations of solutions are generated for each test problem (instance) and used as initial populations. Each algorithm is tested with 30 independent trials from the 30 initial populations, respectively. The optimisation stop criteria for real-value problems and CVRPTW follow the same setting used in \cite{10.1145/3321707.3321813,9852781}. respectively,
The quality of the best solution in the final population is taken as the performance measurement. 

\paragraph{Compared methods}
Two state-of-the-art methods with state-based AOS of each domain, namely DE-DDQN \cite{10.1145/3321707.3321813} for real-value optimisation and DQN-GSF \cite{9852781} for CVRPTW, are compared. The state features designed in DE-DDQN and DQN-GSF allow for different problems (instances) to share the same state space, making learning from experiences of different problems (instances) applicable. DE-DDQN and DQN-GSF differ not only in the model used in state-based AOS methods (i.e., DDQN and DQN), but also in reward functions, state features and other detailed settings. Readers are referred to \cite{10.1145/3321707.3321813} and \cite{9852781} for details about the two state-based AOS methods and training. 
Experiments on real-value problems are conducted based on the source code with the trained policy of DE-DDQN provided in \cite{10.1145/3321707.3321813}. Thus, original algorithm settings are kept and the model is directly taken from \cite{10.1145/3321707.3321813}. For experiments on CVRPTW, we implement the algorithm of DQN-GSF, keeping the settings as described in \cite{9852781} as no source code or model is available.
Both offline and online training of the state-based AOS model follows the same setting as the training process presented in the original study \cite{10.1145/3321707.3321813,9852781}.
Testing runs are conducted independently. For either real-value problems or CVRPTW, the online model updating in each run starts from the identical model offline-trained. 

\paragraph{Implementation of hybrid framework}
As the focus of this paper is not to develop new stateless AOS methods, but to leverage the performance of AOS through offline and online experience learning, DE-DDQN and DQN-GSF are used directly as the meta-heuristic and the state-based AOS module in the framework. \rc{All detailed components, e.g., reward functions, state features, and meta-heuristics together with their parameters, of DE-DDQN \cite{10.1145/3321707.3321813} and DQN-GSF \cite{9852781} are used without changes. } The widely used OSR and CA methods, adaptive pursuit \cite{10.1145/1068009.1068251} and fitness improvement rate \cite{6410018}, compose the stateless AOS module in the hybrid framework.  
According to adaptive pursuit~\cite{10.1145/1068009.1068251}, the probability of selecting operator $i$ among $K$ operators at each iteration is
\begin{equation}
P_{i} \leftarrow \begin{cases} \beta P_{max} + (1-\beta)P_{i}, &  \text{ if } i=\arg\max_j{Q_{j}},\\
  \beta \frac{1-P_{max}}{K-1} + (1-\beta)P_{i}, & \text{ otherwise, }
\end{cases}
\label{equ:ap1}
\end{equation}
where $Q_{i} \leftarrow \alpha  c_i + (1-\alpha) Q_{i}$ and is initialised as 0. The credit assigned to operator $i$ according to the fitness improvement rate~\cite{6410018} is $c_i=\max \left\{0, \frac{f(x)-f(x')}{f(x)} \right\}$, where $f(x)$ and $f(x')$ are the objective values of the original solution and generated new solution, respectively. $P_{i}$ for each operator $i$ is initialised as $\frac{1}{K}$. 
$P_{max}=0.85$, $\alpha =0.01$, and $\beta=0.01$ based on the rule of thumb.
The upper and lower bounds of module selection probability are set as $p_u=0.5$ and $p_l=0.1$, respectively.
The state-based AOS module is selected with a higher probability as it learns from rich offline experiences.

\subsection{Effectiveness of the Hybrid AOS Framework\label{sec:hy_test}}

The hybrid framework is expected to improve the optimisation performance of meta-heuristics without relying on the similarity between offline training problems and ongoing solving (test) problems. Therefore it is expected to perform well in both scenarios that the test problems (instances) own relatively different or similar characteristics from the offline training problems. 
The state-based AOS model of the hybrid framework, DE-DDQN, DQN-GSF and their variants are firstly trained offline on the training problems as described in Section \ref{sec:experiment_setting}. 
Then, we conduct a comparison on the test problems as shown in Tables \ref{tab:real_comparison} and \ref{tab:cvrptw_comparison}. Each cell lists the number of problems that the method of the row is statistically better/comparable/worse than the method of the column according to Wilcoxon signed-rank statistic test with $p<0.05$. Appendix provides detailed results on each problem.

\begin{table*}[htbp]
\centering
\caption{Test performance on 138 real-value optimisation problems from CEC2017 \cite{wu2017problem} and BBOB2023 \cite{nikolaus_hansen_2019_2594848}. Each cell represents the number of problems that the method of the row is statistically better/comparable/worse than the method of the column according to Wilcoxon signed-rank statistic test with $p<0.05$.}
\label{tab:real_comparison}
\begin{tabular}{l|ccccccccc}
\toprule
\multicolumn{1}{c|}{(+/$\approx$/-)}  & HF-NU  & HF-NA(0.5)       & HF-NA(0.3)       & HF-NA(0.1)       & DE-DDQN      & DE-DDQN-U    & SL   & Random  \\
\midrule
HF (Ours)         & 23/113/2         & 5/114/19         & 1/131/6          & 17/117/4         & 36/95/7          & 25/110/3         & 42/94/2          & 79/58/1 \\
HF-NU      & $-$ & 14/103/21        & 5/109/24         & 5/116/17         & 31/99/8          & 10/121/7         & 40/88/10         & 80/57/1 \\
HF-NA(0.5) & $-$ & $-$ & 16/118/4         & 21/107/10        & 39/81/18         & 27/101/10        & 43/92/3          & 79/57/2 \\
HF-NA(0.3) & $-$ & $-$ & $-$ & 21/115/2         & 39/92/7          & 28/107/3         & 45/92/1          & 87/49/2 \\
HF-NA(0.1) & $-$ & $-$ & $-$ & $-$ & 34/95/9          & 22/115/1         & 38/95/5          & 80/56/2 \\
DE-DDQN         & $-$ & $-$ & $-$ & $-$ & $-$ & 6/103/29         & 32/84/22         & 68/69/1 \\
DE-DDQN-U       & $-$ & $-$ & $-$ & $-$ & $-$ & $-$ & 38/91/9          & 79/57/2 \\
SL         & $-$ & $-$ & $-$ & $-$ & $-$ & $-$ & $-$ & 56/78/4  
 \\
\bottomrule
\end{tabular}
\end{table*}

\begin{table*}[htbp]
\centering
\caption{Test performance on 33 CVRPTW C-type and RC-type instances \cite{10.1287/opre.35.2.254}.} 
\label{tab:cvrptw_comparison}
\begin{tabular}{l|ccccccccc}
\toprule
\multicolumn{1}{c|}{(+/$\approx$/-)}  & HF-NU& HF-NA(0.5) & HF-NA(0.3)   & HF-NA(0.1)   & DQN-GSF     & DQN-GSF-U & SL & Random \\
\midrule
HF (Ours)         & 13/15/5          & 21/7/5           & 15/18/0          & 26/2/5           & 21/7/5           & 18/10/5          & 23/3/7           & 23/3/7  \\
HF-NU      & $-$ & 7/13/13          & 5/16/12          & 8/19/6           & 33/0/0           & 1/15/17          & 2/13/18          & 2/13/18 \\
HF-NA(0.5) & $-$ & $-$ & 7/13/13          & 25/7/1           & 16/13/4          & 8/12/13          & 20/5/8           & 20/5/8  \\
HF-NA(0.3) & $-$ & $-$ & $-$ & 26/2/5           & 19/9/5           & 13/15/5          & 17/8/8           & 20/5/8  \\
HF-NA(0.1) & $-$ & $-$ & $-$ & $-$ & 13/16/4          & 0/5/28           & 4/5/24           & 4/9/20  \\
DQN-GSF        & $-$ & $-$ & $-$ & $-$ & $-$ & 0/11/22          & 1/10/22          & 1/10/22 \\
DQN-GSF-U       & $-$ & $-$ & $-$ & $-$ & $-$ & $-$ & 14/12/7          & 13/15/5 \\
SL         & $-$ & $-$ & $-$ & $-$ & $-$ & $-$ & $-$ & 2/30/1  \\
\bottomrule
\end{tabular}
\end{table*}

\begin{figure}[htbp]
  \centering
  \subfigure[CEC2017-F13 30-D]{\includegraphics[width=0.48\linewidth]{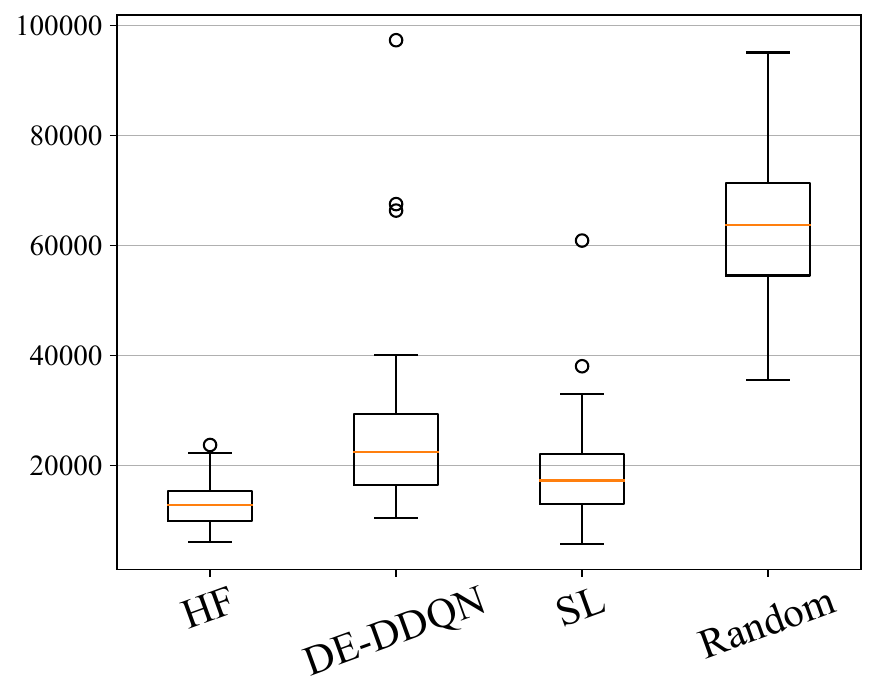}}
  \subfigure[BBOB2023-F17 10-D]{\includegraphics[width=0.48\linewidth]{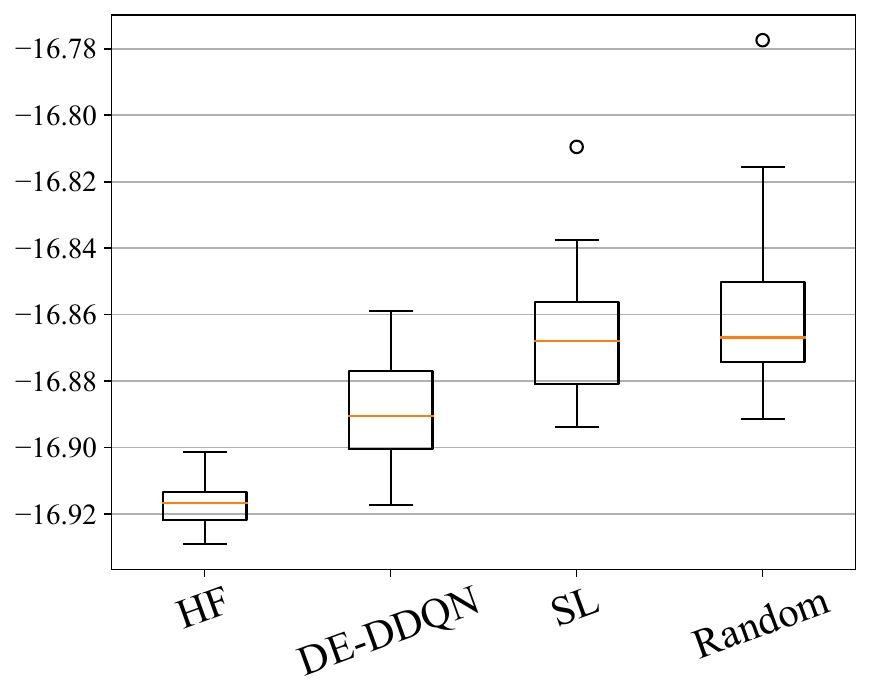}}
  \subfigure[RC106]{\includegraphics[width=0.48\linewidth]{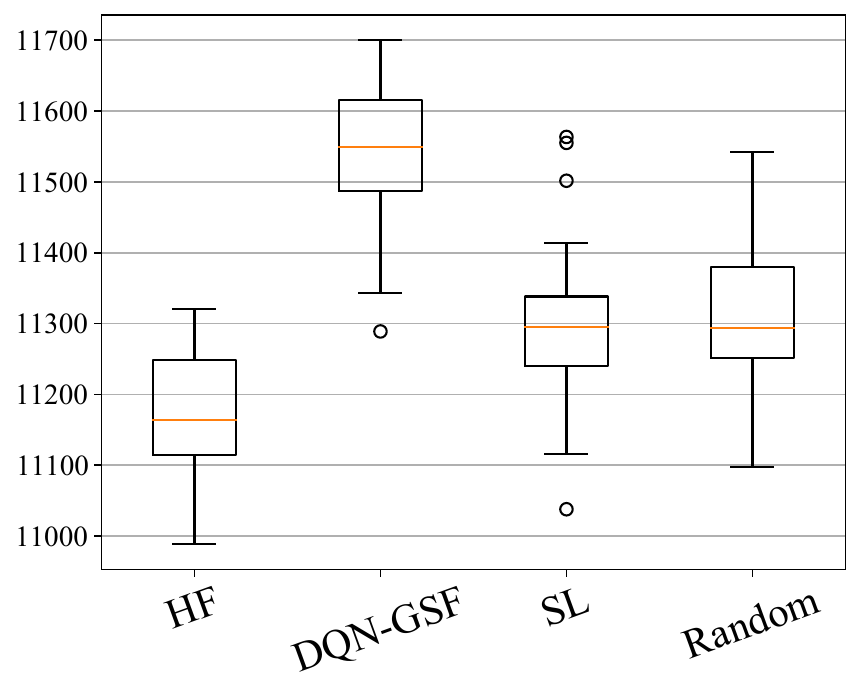}}
  \subfigure[C107]{\includegraphics[width=0.48\linewidth]{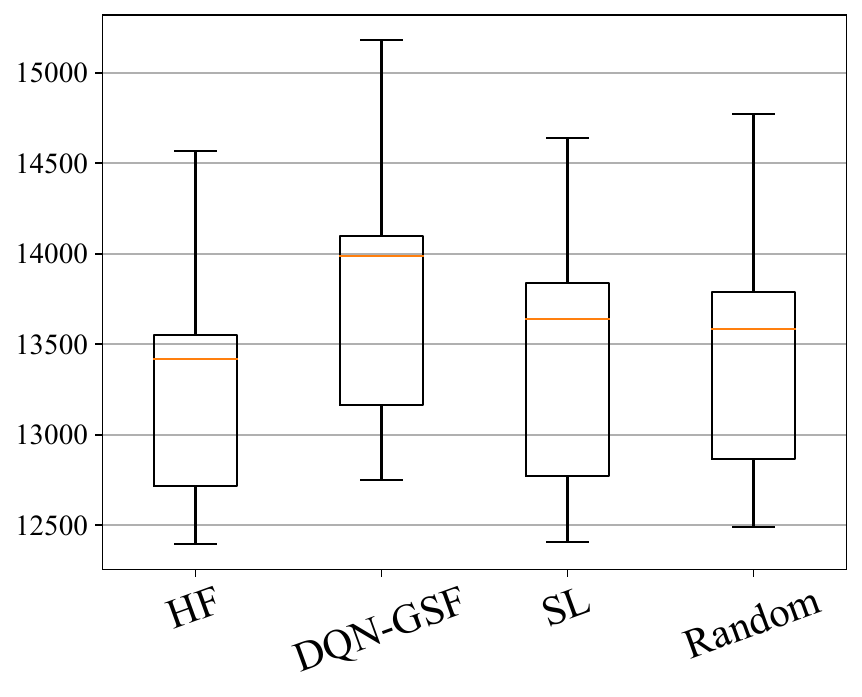}}
  \caption{Average objective value and standard deviation of solutions obtained on two real-value problems (top) and two Solomon CVRPTW instances (bottom).}
  \label{fig:box_fig}
\end{figure}

\paragraph{Performance on test set}According to Table \ref{tab:real_comparison}, for solving real-value problems, HF significantly outperforms DE-DDQN on 36 problems, and is significantly worse than DE-DDQN on only 7 problems. For solving CVRPTW (cf. Table \ref{tab:cvrptw_comparison}), our HF is significantly better and worse than DE-DDQN on 21 and 5 test instances, respectively. 
As illustrative examples, Fig. \ref{fig:box_fig} demonstrates solutions' objective values obtained in solving
CEC2017-F13 of dimension 30, BBOB2023-F17 of dimension 10, Solomon RC106 and C107 instances. 
Learning from only offline experiences, the quality of solutions found by DE-DDQN and DQN-GSF is lower than the ones found by SL, which learns from only online experiences, on three of the four examples. On RC106 and C107, DQN-GSF is even inferior to Random. Our HF, which combines the two AOS modules to learn offline and online experiences is effective in solving test problems (instances) that 
are relatively different to the training ones.
\paragraph{Performance on training set}We also compare the algorithms on training problems in Tables \ref{tab:test_compare_real} and \ref{tab:test_compare_CVRPTW}.

HF performs significantly better than DE-DDQN on 6 real-value problems and significantly worse performance on only 2 problems. 
On the training instance of CVRPTW, the solutions obtained by HF are also significantly better than the one of DQN-GSF, as demonstrated in Table \ref{tab:test_compare_CVRPTW}. 
This observation suggests that the stateless AOS module and decision policy also contribute to the performance of solving training problems.

\begin{table}[htbp]
\caption{Training performance on 32 real-value optimisation problems from CEC2005 \cite{suganthan2005problem}. \label{tab:test_compare_real}}
 \setlength\tabcolsep{1.8pt} {
\begin{tabular}{c|ccc}
\toprule
(+/$\approx$/-)      & DE-DDQN  & SL   & Random  \\
\midrule
HF (Ours)  & 6/24/2 & 9/22/1  & 12/20/0\\
DE-DDQN  & $-$ & 16/15/1 & 12/18/2\\
SL & $-$ & $-$ & 8/22/2 \\
\bottomrule
\end{tabular}}
\end{table}

\begin{table}[htbp]
\caption{Training performance on CVRPTW R101 instance \cite{10.1287/opre.35.2.254}. Average objective value and standard deviation of output solutions are reported. \label{tab:test_compare_CVRPTW}}
\begin{tabular}{c|cccc}
\toprule
  Method & HF (Ours)   & DQN-GSF & SL & Random   \\
\midrule
Avg.  & 14095.20  & 14216.55 & 14248.01 & 14271.41\\
Std.  & 571.51    & 612.16   & 627.90   & 601.32\\
\bottomrule
\end{tabular}
\end{table}

In summary, experiment results indicate that the proposed hybrid framework outperforms state-of-the-art state-based and stateless AOS methods on both seen and unseen problems. The proposed hybrid framework is capable of learning both offline and online experiences and selecting promising operators. 

\subsection{Ablation Study\label{sec:online_experiment}}

The hybrid framework combines the stateless AOS and state-based AOS modules to utilise their strengths.
In the proposed framework, there are two relatively independent online experience learning processes, which are the updating of the state-based AOS model (line \ref{line:SBupdate} of Algorithm \ref{algo:framework}) and the updating of the recording of stateless AOS (line \ref{line:SLupdate} of Algorithm \ref{algo:framework}). Both modules are expected to provide instructive knowledge during the optimisation process of the current problem. To verify their contributions, the two online learning processes in those two modules are tested independently, results of which are also provided in Tables \ref{tab:real_comparison} and \ref{tab:cvrptw_comparison}. 
Four groups of algorithms are compared, including (i) DE-DDQN, DQN-GSF and SL; (ii) DE-DDQN and DQN-GSF with online updating, denoted as DE-DDQN-U and DQN-GSF-U respectively; (iii) a variant of the hybrid framework by removing the online updating, denoted as HF-NU;
and (iv) our hybrid framework.
Its notable that group (i) refers to using state-based or stateless AOS alone.

\paragraph{Leveraging Performance through Online Learning}
According to Tables \ref{tab:real_comparison} and \ref{tab:cvrptw_comparison}, 
DE-DDQN-U achieves significantly better or comparable performance than DE-DDQN on 132 real-value problems and worse performance on only 6 problems. For solving CVRPTW, DQN-GSF-U significantly outperforms DQN-GSF on 22 instances, and achieves comparable performance on 11 instances. DQN-GSF never outperforms DQN-GSF-U on CVRPTW. HF also achieves a similar superiority to HF-NU.
This indicates that updating the model with online experiences improves the optimisation performance.

\paragraph{Cooperation of two AOS modules}Furthermore, HF-NU is significantly better than DQN-GSF on all the 33 CVRPTW test instances, while being better than or comparable with DE-DDQN on 130 real-value problems, indicating that using both state-based and stateless AOS modules is better than using the state-based module alone. However, HF-NU does not show outstanding performance compared to SL, a stateless AOS method, while HF does. This observation indicates that learning new experiences online is necessary.

\subsection{Dynamic Adjustment Helps \label{sec:weight_experiment}}

The two AOS modules have complementary abilities in online learning and operator selection. The decision policy is the key component in the framework to balance the two AOS modules for effective collaboration. To verify the effectiveness of the proposed decision policy, we compare the framework with different probability settings on all test problems with the fixed value of $p$ as $p=0.5$, $p=0.3$ and $p=0.1$, denoted as HF-NA(0.5), HF-NA(0.3) and HF-NA(0.1), respectively. Results are reported in Tables \ref{tab:real_comparison} and \ref{tab:cvrptw_comparison} as well.

Results on the real-value problems imply that a higher probability 
of using the operator selected by 
the stateless AOS module leads to slightly better overall performance, while results on CVRPTW show that HF-NA(0.3) performs overall better than HF-NA(0.5) and HF-NA(0.1).
Though HF is surpassed by HF-NA(0.5) on 19 real-value problems, it shows superior performance on CVRPTW instances, given its design rooted in conclusions drawn from the study of combinatorial optimisation problems \cite{10.1007/978-3-031-20862-1_41}. These observations suggest that HF with dynamic adjustment of $p$ is more general than using a fixed selection probability.
The proposed HF is designed to efficiently tackle problems without depending on prior knowledge about their characteristics.
Generally, incorporating the dynamic adjustment of decision policy enhances the framework's potential to solve new problems.

\section{Conclusions\label{sec:conclusion}}

This paper studies adaptive operator selection (AOS) for meta-heuristics by learning from both offline and online optimisation experiences.
A hybrid AOS framework is proposed. In the framework, a state-based AOS module is employed to learn firstly from offline experiences collected in previously solved problems, and then from online experiences collected in real-time while solving the current problem. A stateless AOS module is embedded into the framework to learn the online experiences from scratch simultaneously with the state-based AOS model. 
A decision policy adjusting mechanism is applied in the framework to balance the two modules.

Through extensive experiments on real-value optimisation problems and a challenging constrained combinatorial optimisation problem, we verify that the proposed hybrid framework outperforms state-of-the-art AOS methods, including DE-DDQN \cite{10.1145/3321707.3321813}, DQN-GSF \cite{9852781} and stateless AOS method \cite{10.1145/1068009.1068251}. In all the scenarios regardless of whether the to-be-solved problems are relatively different from or identical to past ones, the hybrid framework achieves the best performance among all the compared methods.

\rc{In future work, we will investigate the potential ways to improve the framework, e.g., adjusting the decision policy. Furthermore, we will analyse the features of problems that each AOS method tested performs better for better understanding the behaviour of operators and AOS methods. } Future work will also explore the framework's potential in a wider range of problem and meta-heuristic fields.

\newpage
\bibliographystyle{ACM-Reference-Format}
\bibliography{main}


\clearpage
\newpage

\appendix
\setcounter{table}{0} 
\setcounter{figure}{0}
\setcounter{section}{0}
\setcounter{equation}{0}
\renewcommand{\thetable}{S-\arabic{table}}
\renewcommand{\thefigure}{S-\arabic{figure}}
\renewcommand{\thesection}{S-\arabic{section}}
\renewcommand{\theequation}{S-\arabic{equation}}

\section*{Appendix: Detailed Experiment Results}

Tables \ref{tab:cec2017d10}, \ref{tab:cec2017d30}, and \ref{tab:cec2017d50} present the average and standard deviation of 30 output solutions' objective values for each method on the tested CEC2017 problems. Results for BBOB2023 problems and CVRPTW instances are also provided in Tables \ref{tab:bbob2023} and \ref{tab:C_RC}. The comparison results, as discussed in Section \ref{sec:hy_test}, are based on the listed experiment results.
In each table, a cell is marked with "+" if the corresponding 30 objective values are significantly larger than the objective value obtained by the proposed hybrid framework (HF), indicating that the method's performance on this problem (instance) is significantly worse than HF. Conversely, if a cell is marked with "-", it indicates that on this problem (instance), the method obtains significantly smaller objective values and exhibits better performance than HF.

\begin{table*}[hb]
    \centering
    \caption{Output solution's average objective value and standard deviation from 30 runs on CEC2017 benchmark problems in 10-dimension. "+" or "-" indicate the objective values are significantly larger (worse) or smaller (better) than objective values obtained by HF, respectively.}
    \begin{adjustbox}{width=\textwidth,center}
     \setlength\tabcolsep{1pt}{
    \begin{tabular}{c|ccccccccc}
\toprule
    & HF (Ours)                  & HF-NU        & HF-NA(0.5)         & HF-NA(0.3)         & HF-NA(0.1)       & DE-DDQN     & DE-DDQN-U             & SL                 & Random                \\
    \midrule
F1  & 886.42(6.8e+02)   & 858.26(7.7e+02)   & 785.44(1.1e+03)   & 1129.90(8.3e+02)  & 2215.69(1.9e+03)+ & 8663.60(4.9e+03)+ & 4573.76(2.4e+03)+ & 289.21(2.3e+02)-  & 87133.43(4.2e+04)+ \\
F2  & 200.00(3.38e-04)  & 200.00(0.0017)+   & 200.00(0.014)     & 200.00(2.96e-04)  & 200.00(0.0016)+   & 200.00(0.0021)+   & 200.00(0.0014)+   & 200.00(0.002)+    & 200.07(0.057)+     \\
F3  & 300.00(5.29e-04)  & 300.01(0.0035)+   & 300.00(2.39e-04)- & 300.00(5.26e-04)  & 300.00(0.0015)+   & 300.01(0.0085)+   & 300.01(0.0036)+   & 300.00(0.0019)+   & 300.12(0.045)+     \\
F4  & 401.76(0.86)      & 401.77(0.74)      & 402.13(0.96)      & 401.78(0.99)      & 402.01(1)         & 401.28(0.69)-     & 401.69(0.94)      & 402.08(0.89)      & 403.09(0.82)+      \\
F5  & 535.38(4.8)       & 534.42(6.4)       & 531.77(6.3)-      & 535.52(5.9)       & 534.03(4.7)       & 537.89(4.4)+      & 533.74(5.2)       & 534.48(5.1)       & 535.82(6.4)        \\
F6  & 600.38(0.13)      & 600.51(0.16)+     & 600.24(0.077)-    & 600.31(0.13)-     & 600.45(0.18)      & 600.71(0.22)+     & 600.52(0.2)+      & 600.37(0.27)      & 601.27(0.32)+      \\
F7  & 745.89(4.7)       & 746.68(6.4)       & 746.91(4.5)       & 748.11(5)         & 746.60(6.2)       & 748.62(6.4)+      & 747.31(5.3)       & 746.21(5.1)       & 749.82(5.9)+       \\
F8  & 835.58(5.8)       & 835.16(5.9)       & 833.80(4)         & 836.39(4.7)       & 835.94(4.6)       & 836.03(4.8)       & 835.13(5.7)       & 835.15(4.8)       & 839.32(4.8)+       \\
F9  & 900.01(0.013)     & 900.07(0.043)+    & 900.01(0.0078)-   & 900.01(0.011)     & 900.03(0.02)+     & 900.17(0.14)+     & 900.08(0.044)+    & 900.03(0.023)+    & 900.41(0.14)+      \\
F10 & 2460.06(2.8e+02)  & 2455.58(2.3e+02)  & 2491.91(2.9e+02)  & 2504.98(2.3e+02)  & 2516.94(3e+02)    & 2489.28(2e+02)    & 2529.06(2.4e+02)  & 2516.32(2.7e+02)  & 2501.96(2.4e+02)   \\
F11 & 1108.28(1.9)      & 1108.34(1.6)      & 1107.82(1.6)      & 1108.66(1.7)      & 1108.57(1.6)      & 1109.15(2.3)      & 1108.63(2)        & 1109.16(2.8)      & 1110.35(1.6)+      \\
F12 & 1476.79(1.1e+02)  & 1495.74(1.1e+02)  & 1469.96(1.1e+02)  & 1477.55(77)       & 1456.45(1.3e+02)  & 1524.39(96)       & 1527.82(1.3e+02)  & 1454.18(1.1e+02)  & 1641.82(1.1e+02)+  \\
F13 & 1314.92(2.7)      & 1315.58(4.1)      & 1314.82(4.2)      & 1314.56(4.1)      & 1312.87(3.3)-     & 1316.10(3.6)      & 1314.62(4.2)      & 1315.81(4.4)      & 1319.07(5.2)+      \\
F14 & 1429.05(2.7)      & 1429.32(2.7)      & 1428.48(2.8)      & 1428.93(2.7)      & 1428.65(4.2)      & 1430.04(2.4)      & 1429.38(2.5)      & 1428.68(3.6)      & 1430.67(3.8)+      \\
F15 & 1504.10(1)        & 1504.32(1)        & 1503.70(1)        & 1504.07(0.96)     & 1504.23(1)        & 1504.75(0.86)+    & 1504.41(1.1)      & 1504.04(1.1)      & 1505.32(1.2)+      \\
F16 & 1689.22(50)       & 1670.53(47)       & 1674.75(33)       & 1673.99(39)       & 1661.78(42)-      & 1669.63(44)       & 1677.96(38)       & 1675.98(41)       & 1698.22(43)        \\
F17 & 1768.85(13)       & 1774.65(20)       & 1766.95(14)       & 1771.58(16)       & 1771.21(17)       & 1769.36(17)       & 1770.28(13)       & 1766.14(17)       & 1776.42(16)        \\
F18 & 1819.47(3.9)      & 1819.19(3.9)      & 1816.75(4.9)-     & 1816.83(5.1)      & 1819.66(4.4)      & 1819.68(3.9)      & 1818.07(4.4)-     & 1818.51(4.1)      & 1820.65(3.9)       \\
F19 & 1902.88(0.65)     & 1903.04(0.78)     & 1902.97(0.85)     & 1902.92(0.74)     & 1903.15(0.71)     & 1903.20(0.67)     & 1902.88(0.66)     & 1903.14(0.65)     & 1903.91(0.58)+     \\
F20 & 2095.76(26)       & 2097.35(21)       & 2107.63(28)+      & 2097.75(27)       & 2093.90(26)       & 2099.84(22)       & 2098.39(20)       & 2106.05(33)       & 2124.65(22)+       \\
F21 & 2314.08(51)       & 2308.20(54)       & 2322.65(40)       & 2313.40(51)       & 2305.68(58)       & 2314.26(51)       & 2308.86(55)       & 2290.72(64)       & 2306.22(58)        \\
F22 & 2295.52(25)       & 2296.40(24)+      & 2298.66(17)       & 2298.98(17)       & 2298.92(18)       & 2295.33(24)-      & 2299.59(16)       & 2302.04(0.52)     & 2298.42(19)+       \\
F23 & 2637.70(4.9)      & 2636.23(5)        & 2636.12(5.8)      & 2636.59(6)        & 2636.26(4)        & 2635.27(6)        & 2635.73(6.3)      & 2637.88(5)        & 2639.22(6.4)       \\
F24 & 2728.18(90)       & 2738.76(80)+      & 2745.50(66)       & 2736.20(79)       & 2745.54(66)       & 2728.79(90)       & 2738.64(80)       & 2754.80(48)       & 2748.39(66)+       \\
F25 & 2912.03(21)       & 2918.26(23)       & 2913.50(22)       & 2913.67(22)       & 2913.45(22)       & 2913.75(22)       & 2908.97(19)       & 2918.18(23)       & 2908.82(19)        \\
F26 & 2900.05(0.015)    & 2900.14(0.029)+   & 2900.03(0.0094)-  & 2900.04(0.014)    & 2901.62(8.5)      & 2900.13(0.043)+   & 2900.10(0.038)+   & 2900.16(0.19)+    & 2900.52(0.18)+     \\
F27 & 3090.32(1.8)      & 3090.65(1.4)      & 3091.23(2.3)      & 3091.09(2.7)      & 3090.93(2.2)      & 3090.50(1.8)      & 3090.62(2.1)      & 3091.23(2)        & 3091.05(1.2)+      \\
F28 & 3234.69(1.5e+02)  & 3207.62(1.4e+02)  & 3224.16(1.4e+02)  & 3231.08(1.4e+02)  & 3201.76(1.4e+02)  & 3196.78(1.3e+02)  & 3214.08(1.4e+02)  & 3179.64(1.2e+02)  & 3153.30(1e+02)     \\
F29 & 3188.47(19)       & 3202.77(30)+      & 3193.76(24)       & 3194.41(26)       & 3187.01(21)       & 3196.34(25)       & 3198.50(25)+      & 3197.46(27)       & 3205.15(23)+       \\
F30 & 1.5e+05(3.68e+05) & 3.6e+05(4.88e+05) & 2.4e+05(4.78e+05) & 1.4e+05(3.61e+05) & 3.8e+05(5.07e+05) & 2.2e+05(4.52e+05) & 2.4e+05(4.79e+05) & 1.6e+05(3.51e+05) & 1e+05(3.10e+05)-    \\
\bottomrule
    \end{tabular}
    }
    \end{adjustbox}
    \label{tab:cec2017d10}
\end{table*}

\begin{table*}[h]
    \centering
    \caption{Output solution's average objective value and standard deviation from 30 runs on CEC2017 benchmark problems in 30-dimension. "+" or "-" indicate the objective values are significantly larger (worse) or smaller (better) than objective values obtained by HF, respectively.}
    \begin{adjustbox}{width=\textwidth,center}
     \setlength\tabcolsep{1pt}{
    \begin{tabular}{c|ccccccccc}
\toprule
  & HF (Ours)                  & HF-NU        & HF-NA(0.5)         & HF-NA(0.3)         & HF-NA(0.1)       & DE-DDQN    & DE-DDQN-U             & SL                 & Random                \\
    \midrule
F1  & 3.9e+10(2.04e+10) & 5e+10(3.05e+10)   & 4.3e+10(2.12e+10) & 5.1e+10(3.88e+10) & 4.2e+10(2.17e+10) & 3.1e+10(1.56e+10)  & 3.5e+10(1.82e+10)  & 5e+10(2.38e+10)    & 4.8e+10(2.35e+10)  \\
F2  & 1.7e+20(7.76e+20) & 1e+24(4.02e+24)   & 2.4e+20(5.92e+20) & 1.6e+22(8.55e+22) & 9.9e+21(4.62e+22) & 6.5e+19(1.99e+20)  & 1.2e+21(4.80e+21)  & 1e+25(3.73e+25)+   & 1.5e+23(6.68e+23)+ \\
F3  & 14330.00(6e+03)   & 13164.47(6.6e+03) & 11715.48(5.5e+03) & 14007.74(5.6e+03) & 13851.89(6.7e+03) & 15431.49(5.9e+03)  & 13614.78(6.9e+03)  & 18906.76(8.3e+03)+ & 17094.71(7.6e+03)  \\
F4  & 742.12(2.1e+02)   & 714.56(1.4e+02)   & 839.64(1.8e+02)+  & 760.59(2.3e+02)   & 806.47(3.8e+02)   & 727.99(1.8e+02)    & 720.17(1.9e+02)    & 809.28(2.4e+02)    & 869.90(2.2e+02)+   \\
F5  & 744.02(15)        & 740.80(18)        & 739.37(19)        & 741.27(18)        & 740.16(15)        & 746.43(16)         & 740.44(19)         & 755.57(24)+        & 751.55(20)         \\
F6  & 636.25(6.6)       & 630.78(8.4)-      & 636.77(7.9)       & 635.05(9.6)       & 635.33(11)        & 636.55(10)         & 633.87(10)         & 635.53(12)         & 638.73(9.2)        \\
F7  & 996.48(29)        & 998.71(26)        & 989.48(24)        & 990.28(26)        & 998.58(31)        & 1013.36(29)+       & 1004.15(26)        & 1008.11(28)        & 1009.10(30)        \\
F8  & 1025.98(18)       & 1028.94(17)       & 1028.93(15)       & 1029.52(18)       & 1032.93(13)+      & 1040.69(19)+       & 1030.44(18)        & 1028.48(17)        & 1037.96(18)+       \\
F9  & 2411.32(7e+02)    & 2402.58(6.4e+02)  & 2488.44(6.4e+02)  & 2409.12(6.8e+02)  & 2337.56(6.5e+02)  & 2965.13(9.6e+02)+  & 2215.79(6.3e+02)   & 2612.33(7.3e+02)   & 2945.10(9.2e+02)+  \\
F10 & 8932.64(3.2e+02)  & 8869.70(3.2e+02)  & 8966.84(2.2e+02)  & 8810.39(3.3e+02)  & 8895.44(3.2e+02)  & 8810.60(4.1e+02)   & 8914.25(2.7e+02)   & 8967.22(2.8e+02)   & 8972.48(2.5e+02)   \\
F11 & 1293.56(48)       & 1294.87(44)       & 1305.87(60)       & 1293.74(48)       & 1318.03(54)+      & 1319.76(43)        & 1308.74(57)        & 1314.45(68)        & 1326.22(68)        \\
F12 & 6.1e+06(7.13e+06) & 4.7e+06(5.47e+06) & 4.4e+06(8.23e+06) & 6.2e+06(1.14e+07) & 4.8e+06(8.20e+06) & 7.7e+06(1.47e+07)  & 8.8e+06(1.91e+07)  & 5.8e+06(6.38e+06)  & 2.6e+07(2.60e+07)+ \\
F13 & 13102.02(4.5e+03) & 13092.86(6.4e+03) & 15483.17(8.3e+03) & 12363.21(4e+03)   & 16607.83(8.3e+03) & 27053.09(1.9e+04)+ & 19254.07(1.3e+04)+ & 19432.44(1.1e+04)+ & 63486.81(1.4e+04)+ \\
F14 & 1606.22(33)       & 1608.18(43)       & 1607.02(43)       & 1595.71(41)       & 1614.37(38)       & 1628.95(52)        & 1616.02(47)        & 1610.92(29)        & 1608.37(25)        \\
F15 & 2069.12(2e+02)    & 2056.84(1.8e+02)  & 2207.14(3.4e+02)  & 2148.64(1.7e+02)+ & 2270.00(6.5e+02)+ & 2211.35(3.6e+02)   & 2150.27(2.7e+02)   & 2457.69(5.3e+02)+  & 2792.45(6.3e+02)+  \\
F16 & 3454.60(1.8e+02)  & 3581.64(2.4e+02)+ & 3474.67(2.3e+02)  & 3465.05(2.2e+02)  & 3488.68(2.2e+02)  & 3543.55(2.4e+02)   & 3489.14(2.3e+02)   & 3473.59(1.8e+02)   & 3555.48(2.1e+02)   \\
F17 & 2445.26(1.3e+02)  & 2468.71(1.6e+02)  & 2478.75(1.5e+02)  & 2470.11(1.5e+02)  & 2456.78(1.6e+02)  & 2540.24(1.2e+02)+  & 2522.52(1.4e+02)   & 2513.22(1.4e+02)+  & 2541.26(1.3e+02)+  \\
F18 & 2131.19(92)       & 2124.70(1.2e+02)  & 2184.23(2.7e+02)  & 2101.14(1.3e+02)  & 2110.18(98)       & 2210.46(1.5e+02)+  & 2173.66(1.2e+02)   & 2232.81(1.3e+02)+  & 2519.81(2e+02)+    \\
F19 & 2080.01(57)       & 2081.91(76)       & 2054.98(84)       & 2097.52(1.2e+02)  & 2097.86(87)       & 2151.02(1.1e+02)+  & 2112.62(72)        & 2105.26(1.1e+02)   & 2577.77(3.3e+02)+  \\
F20 & 2853.48(1.4e+02)  & 2878.74(1.4e+02)  & 2888.58(1.2e+02)  & 2879.18(1.4e+02)  & 2862.74(1.4e+02)  & 2852.99(1.5e+02)   & 2885.43(1.3e+02)   & 2905.05(1.3e+02)   & 2937.67(1.4e+02)+  \\
F21 & 2528.73(18)       & 2524.61(24)       & 2529.43(21)       & 2526.41(19)       & 2534.68(18)       & 2538.84(23)        & 2527.84(19)        & 2529.32(15)        & 2545.91(21)+       \\
F22 & 3974.15(2.8e+03)  & 4502.41(3.2e+03)  & 5165.85(3.3e+03)  & 4288.20(3e+03)    & 4434.47(3.1e+03)  & 4398.90(3.1e+03)   & 3926.08(2.8e+03)   & 3681.60(2.1e+03)   & 3824.20(2.3e+03)   \\
F23 & 2972.88(51)       & 2944.82(42)       & 2959.32(48)       & 2937.70(30)-      & 2958.17(52)       & 2948.24(29)-       & 2958.36(43)        & 2961.27(50)        & 2963.10(49)        \\
F24 & 3119.79(36)       & 3139.33(43)       & 3124.53(53)       & 3118.43(46)       & 3143.09(43)       & 3136.28(50)        & 3141.57(53)        & 3134.84(37)        & 3144.59(52)+       \\
F25 & 3026.55(67)       & 3056.60(71)       & 3035.16(51)       & 3038.70(60)       & 2995.92(48)-      & 3014.01(47)        & 3028.26(59)        & 3075.70(81)+       & 3055.71(63)        \\
F26 & 6163.06(8.7e+02)  & 6169.67(7.2e+02)  & 6207.98(7.2e+02)  & 6250.27(4.8e+02)  & 6153.66(8e+02)    & 6363.38(6e+02)     & 6533.16(4e+02)     & 6039.22(1e+03)     & 6473.43(3.2e+02)   \\
F27 & 3255.01(37)       & 3252.53(23)       & 3256.83(28)       & 3255.01(22)       & 3257.79(29)       & 3239.25(21)        & 3245.58(25)        & 3269.26(50)        & 3260.59(21)        \\
F28 & 3399.10(87)       & 3400.40(79)       & 3401.80(98)       & 3381.63(77)       & 3365.16(65)       & 3346.20(61)-       & 3377.99(93)        & 3415.38(99)        & 3412.21(98)        \\
F29 & 4437.22(1.5e+02)  & 4400.69(1.7e+02)  & 4423.93(1.7e+02)  & 4444.72(1.5e+02)  & 4428.61(1.9e+02)  & 4436.20(1.9e+02)   & 4424.01(2e+02)     & 4452.54(1.7e+02)   & 4513.33(1.7e+02)+  \\
F30 & 30132.99(2.1e+04) & 4.6e+04(1.20e+05) & 36076.40(4e+04)   & 32042.41(1.9e+04) & 32207.73(2.5e+04) & 37011.70(2.9e+04)  & 6.5e+04(1.79e+05)  & 2.5e+05(1.05e+06)  & 95068.30(5.8e+04)+\\
\bottomrule
    \end{tabular}
    }
    \end{adjustbox}
    \label{tab:cec2017d30}
\end{table*}

\begin{table*}[h]
    \centering
    \caption{Output solution's average objective value and standard deviation from 30 runs on CEC2017 benchmark problems in 30-dimension. "+" or "-" indicate the objective values are significantly larger (worse) or smaller (better) than objective values obtained by HF, respectively.}
    \begin{adjustbox}{width=\textwidth,center}
     \setlength\tabcolsep{1pt}{
    \begin{tabular}{c|ccccccccc}
\toprule
  & HF (Ours)                  & HF-NU        & HF-NA(0.5)         & HF-NA(0.3)         & HF-NA(0.1)       & DE-DDQN     & DE-DDQN-U             & SL                 & Random                \\
    \midrule
    F1  & 3.9e+11(9.54e+10) & 3.8e+11(8.21e+10)  & 3.9e+11(8.84e+10)  & 3.9e+11(6.59e+10) & 3.8e+11(8.63e+10)  & 3.9e+11(9.60e+10)  & 3.7e+11(8.91e+10)  & 4.1e+11(8.83e+10)  & 3.8e+11(7.70e+10)   \\
F2  & 2.7e+51(1.42e+52) & 1.5e+50(4.83e+50)  & 3.2e+50(1.29e+51)  & 3.6e+52(1.66e+53) & 3e+51(9.95e+51)    & 1.1e+51(5.04e+51)  & 3.2e+49(9.52e+49)  & 5.9e+54(2.19e+55)+ & 3.9e+54(1.99e+55)+  \\
F3  & 74780.88(1.6e+04) & 77901.86(1.9e+04)  & 71911.28(1.3e+04)  & 71350.94(1.9e+04) & 72907.04(1.7e+04)  & 76476.56(2.1e+04)  & 74895.09(1.8e+04)  & 75721.99(1.6e+04)  & 81683.37(1.6e+04)   \\
F4  & 5422.12(2.1e+03)  & 5242.97(2.2e+03)   & 5594.40(1.6e+03)   & 4855.69(1.7e+03)  & 4911.65(1.4e+03)   & 4744.68(1.7e+03)-  & 4838.67(1.5e+03)   & 5818.34(2.3e+03)   & 5573.72(1.9e+03)    \\
F5  & 1005.34(32)       & 999.27(29)         & 1003.04(32)        & 1003.64(27)       & 1001.23(40)        & 1011.68(31)        & 1009.63(32)        & 1013.24(27)        & 1016.89(29)         \\
F6  & 674.44(10)        & 672.78(11)         & 679.40(12)         & 673.22(9.8)       & 677.03(8.5)        & 676.92(12)         & 674.54(8.8)        & 678.91(9.8)        & 677.68(9.8)         \\
F7  & 1476.33(83)       & 1458.24(69)        & 1476.39(68)        & 1470.12(87)       & 1464.58(80)        & 1480.23(84)        & 1464.90(72)        & 1530.09(73)+       & 1500.92(88)         \\
F8  & 1302.50(29)       & 1310.84(36)        & 1300.19(27)        & 1304.25(29)       & 1302.19(42)        & 1322.49(30)+       & 1314.88(35)+       & 1323.65(39)+       & 1315.33(30)+        \\
F9  & 16618.36(3.5e+03) & 16126.25(3.7e+03)  & 16468.22(3.5e+03)  & 16040.34(3.3e+03) & 15899.52(3.8e+03)  & 16332.57(2.8e+03)  & 16453.48(4.1e+03)  & 17867.30(3.1e+03)  & 17572.65(4.8e+03)   \\
F10 & 15376.99(5e+02)   & 15629.94(3.8e+02)+ & 15478.24(4.8e+02)  & 15416.27(3.8e+02) & 15525.85(3.8e+02)  & 15635.61(4.3e+02)  & 15482.53(4e+02)    & 15520.24(4.8e+02)  & 15764.27(4.3e+02)+  \\
F11 & 2301.97(3.8e+02)  & 2497.21(6.3e+02)   & 3001.85(1e+03)+    & 2343.56(4.4e+02)  & 2577.78(6.6e+02)   & 2168.73(3.8e+02)   & 2421.38(5.5e+02)   & 2858.81(8.3e+02)+  & 2578.41(5.2e+02)+   \\
F12 & 6.1e+09(6.58e+09) & 3.9e+09(2.38e+09)  & 5.6e+09(3.99e+09)  & 6e+09(4.58e+09)   & 4.8e+09(4.86e+09)  & 2.7e+09(2.88e+09)- & 3.3e+09(2.10e+09)  & 6.1e+09(4.95e+09)  & 8.3e+09(6.07e+09)   \\
F13 & 4.1e+05(1.88e+05) & 6.2e+05(1.29e+06)  & 3.1e+05(2.26e+05)- & 3e+05(1.26e+05)-  & 7.6e+05(4.03e+05)+ & 4.3e+06(2.74e+06)+ & 1.1e+06(1.01e+06)+ & 2.1e+06(3.93e+06)+ & 1.4e+07(5.82e+06)+  \\
F14 & 1907.06(70)       & 1916.48(98)        & 1888.18(86)        & 1900.64(72)       & 1898.40(59)        & 2001.76(94)+       & 1885.73(62)        & 1923.95(92)        & 2027.67(94)+        \\
F15 & 15937.02(7e+03)   & 17832.48(1e+04)    & 15370.03(8.8e+03)  & 15704.13(8.2e+03) & 21953.64(1.2e+04)+ & 80625.90(7.1e+04)+ & 28507.65(1.3e+04)+ & 17531.40(1e+04)    & 186684.92(7.4e+04)+ \\
F16 & 5355.94(2.1e+02)  & 5245.83(3.4e+02)   & 5279.25(2.5e+02)   & 5327.34(3.6e+02)  & 5207.88(3.4e+02)   & 5298.99(3e+02)     & 5293.61(3.5e+02)   & 5254.58(3.6e+02)   & 5442.06(3.7e+02)    \\
F17 & 4127.56(2.1e+02)  & 4059.04(2.2e+02)   & 4058.82(1.7e+02)   & 4068.30(1.8e+02)  & 4097.15(2.9e+02)   & 4201.93(1.9e+02)   & 4118.18(2.2e+02)   & 4071.38(2e+02)     & 4238.47(1.7e+02)    \\
F18 & 17869.81(1.1e+04) & 16677.51(9.1e+03)  & 25224.05(2.3e+04)  & 22265.01(1.3e+04) & 17203.28(9.4e+03)  & 21082.13(2.1e+04)  & 21372.74(1.3e+04)  & 27674.78(2e+04)+   & 33713.28(1.4e+04)+  \\
F19 & 34967.05(3.6e+04) & 48887.52(6.1e+04)  & 36385.63(3.2e+04)  & 48012.24(4.6e+04) & 6.9e+04(1.63e+05)  & 9.9e+04(1.06e+05)+ & 59667.83(5.9e+04)+ & 1.2e+05(2.02e+05)+ & 2.8e+05(1.92e+05)+  \\
F20 & 4253.81(1.7e+02)  & 4179.53(1.9e+02)   & 4239.72(1.2e+02)   & 4240.15(2.3e+02)  & 4251.95(1.9e+02)   & 4136.93(2e+02)     & 4153.25(1.8e+02)-  & 4183.46(1.8e+02)   & 4259.34(1.5e+02)    \\
F21 & 2835.60(49)       & 2823.79(44)        & 2816.43(42)        & 2836.03(38)       & 2829.19(37)        & 2844.08(41)        & 2822.10(35)        & 2837.52(37)        & 2860.69(44)+        \\
F22 & 17084.89(4.7e+02) & 17107.65(3.3e+02)  & 16976.75(4.3e+02)  & 16960.06(4.4e+02) & 17113.24(3.5e+02)  & 16983.91(4.7e+02)  & 16929.28(4.1e+02)  & 16958.56(4.5e+02)  & 17202.80(3.7e+02)   \\
F23 & 3534.49(98)       & 3564.06(1.1e+02)   & 3519.93(97)        & 3566.09(1.1e+02)  & 3526.30(1.1e+02)   & 3539.00(99)        & 3567.83(1.1e+02)   & 3553.69(1.1e+02)   & 3552.01(96)         \\
F24 & 3766.75(93)       & 3774.79(1e+02)     & 3776.24(1.3e+02)   & 3733.20(1.1e+02)  & 3743.75(1.1e+02)   & 3762.06(1.1e+02)   & 3784.33(1e+02)     & 3766.03(1.3e+02)   & 3839.63(1.4e+02)+   \\
F25 & 5917.76(1.3e+03)  & 5715.22(7.9e+02)   & 6150.18(1e+03)     & 5661.69(9.2e+02)  & 5514.79(6.5e+02)   & 5497.35(1e+03)     & 5824.23(9.2e+02)   & 5926.34(8.9e+02)   & 6083.57(1e+03)      \\
F26 & 11423.91(9.8e+02) & 11009.44(8.1e+02)- & 11540.08(8.3e+02)  & 11487.39(1e+03)   & 11445.42(8.7e+02)  & 11573.82(7.3e+02)  & 11442.27(1.1e+03)  & 11205.71(1.5e+03)  & 11801.07(6.9e+02)   \\
F27 & 3724.14(1.3e+02)  & 3755.73(2.1e+02)   & 3711.43(1.3e+02)   & 3676.99(1.3e+02)  & 3719.95(1.8e+02)   & 3658.38(1.2e+02)   & 3741.43(1.7e+02)   & 3705.34(1.3e+02)   & 3852.54(1.6e+02)+   \\
F28 & 5789.29(7.7e+02)  & 5758.52(8.3e+02)   & 5789.32(8.8e+02)   & 5662.28(5.6e+02)  & 5450.11(8.1e+02)   & 5605.00(7.2e+02)   & 5652.92(6e+02)     & 6187.84(7.5e+02)   & 5604.02(7.2e+02)    \\
F29 & 5993.13(3.7e+02)  & 6010.76(4.2e+02)   & 5908.64(3.8e+02)   & 5983.40(3.7e+02)  & 6071.68(3.9e+02)   & 6083.46(4.1e+02)   & 6037.81(3e+02)     & 6096.34(3.2e+02)   & 6166.76(3e+02)      \\
F30 & 3e+07(1.16e+07)   & 2.6e+07(1.21e+07)  & 3e+07(1.22e+07)    & 2.8e+07(1.04e+07) & 2.5e+07(8.85e+06)  & 2.9e+07(2.44e+07)  & 3e+07(1.11e+07)    & 3.8e+07(1.84e+07)  & 4.5e+07(1.53e+07)+ 
   \\
\bottomrule
    \end{tabular}
    }
    \end{adjustbox}
    \label{tab:cec2017d50}
\end{table*}

\begin{table*}[h]
    \centering
    \caption{Output solution's average objective value and standard deviation from 30 runs on BBOB2023 benchmark problems in 10-dimension and 30-dimension. "+" or "-" indicate the objective values are significantly larger (worse) or smaller (better) than objective values obtained by HF, respectively.}
    \begin{adjustbox}{width=\textwidth,center}
     \setlength\tabcolsep{1pt}{
    \begin{tabular}{c|ccccccccc}
\toprule
  & HF (Ours)                  & HF-NU        & HF-NA(0.5)         & HF-NA(0.3)         & HF-NA(0.1)       & DE-DDQN     & DE-DDQN-U             & SL                 & Random                \\
    \midrule
F1-D10  & 79.48(5.43e-07)   & 79.48(3.26e-06)+  & 79.48(8.15e-08)-   & 79.48(3.08e-07)-  & 79.48(9.10e-07)   & 79.48(3.56e-06)+  & 79.48(1.96e-06)+  & 79.48(1.44e-06)+  & 79.48(1.71e-05)+  \\
F2-D10  & -209.88(7.62e-04) & -209.88(0.0031)+  & -209.88(7.19e-04)- & -209.88(8.22e-04) & -209.88(0.0016)+  & -209.87(0.0041)+  & -209.88(0.0024)+  & -209.85(0.032)+   & -209.80(0.037)+   \\
F3-D10  & -425.89(5.7)      & -425.84(4.4)      & -426.99(4.5)       & -425.78(4.8)      & -425.96(5.9)      & -425.86(6.3)      & -426.72(4.8)      & -424.28(5.2)      & -420.90(4.6)+     \\
F4-D10  & -419.41(5.7)      & -417.26(5.9)      & -420.56(6.8)       & -421.18(9.8)      & -417.81(5.1)      & -415.03(5.4)+     & -419.04(6.2)      & -417.73(6)        & -415.70(6.7)      \\
F5-D10  & -9.09(0.18)       & -8.85(0.52)+      & -8.98(0.55)        & -9.09(0.18)       & -9.11(0.15)       & -9.09(0.16)       & -8.49(1.8)+       & -8.00(0.99)+      & -5.62(2.9)+       \\
F6-D10  & 35.96(0.023)      & 35.97(0.022)      & 35.95(0.015)       & 35.95(0.013)      & 35.96(0.02)       & 35.99(0.043)+     & 35.96(0.036)      & 36.00(0.028)+     & 36.31(0.13)+      \\
F7-D10  & 92.94(2.12e-10)   & 92.94(6.20e-09)+  & 92.94(4.92e-11)-   & 92.94(2.41e-10)   & 92.94(1.69e-09)+  & 92.94(1.77e-08)+  & 92.94(3.86e-09)+  & 92.94(1.14e-09)+  & 92.94(0.0036)+    \\
F8-D10  & 153.37(1.4)       & 153.35(1.2)       & 153.58(1.2)        & 153.30(1.2)       & 153.26(1.3)       & 153.24(1.4)       & 153.56(1.4)       & 153.74(1.4)       & 155.02(0.98)+     \\
F9-D10  & 128.03(1.2)       & 128.12(1.3)       & 127.97(1)          & 127.98(1.4)       & 128.14(1.1)       & 127.76(1.3)       & 128.08(1)         & 128.73(0.95)+     & 129.84(1.1)+      \\
F10-D10 & -54.94(0.002)     & -54.93(0.0045)+   & -54.94(8.13e-04)-  & -54.94(0.0016)    & -54.94(0.0033)+   & -54.92(0.009)+    & -54.93(0.0065)+   & -54.92(0.018)+    & -54.71(0.12)+     \\
F11-D10 & 76.27(6.96e-06)   & 76.27(4.77e-05)+  & 76.27(3.25e-06)-   & 76.27(5.74e-06)   & 76.27(1.72e-05)+  & 76.27(5.40e-05)+  & 76.27(4.83e-05)+  & 76.27(5.22e-05)+  & 76.27(3.73e-04)+  \\
F12-D10 & -619.42(1.2)      & -617.25(6.4)      & -619.45(2.9)       & -618.97(3)        & -617.18(6.4)+     & -605.65(16)+      & -614.35(7.4)+     & -618.74(1.8)      & -545.51(36)+      \\
F13-D10 & 30.04(0.019)      & 30.08(0.026)+     & 30.02(0.015)-      & 30.03(0.02)-      & 30.07(0.04)+      & 30.19(0.081)+     & 30.13(0.05)+      & 30.02(0.015)-     & 30.80(0.17)+      \\
F14-D10 & -52.35(3.76e-06)  & -52.35(2.45e-05)+ & -52.35(9.88e-07)-  & -52.35(2.17e-06)  & -52.35(5.87e-06)+ & -52.35(2.17e-05)+ & -52.35(1.16e-05)+ & -52.35(5.70e-05)+ & -52.35(8.76e-05)+ \\
F15-D10 & 1035.52(5.3)      & 1038.67(4.4)+     & 1033.05(3.8)-      & 1034.23(5.7)      & 1037.83(5.3)      & 1036.87(4.4)      & 1037.12(5)        & 1035.72(5)        & 1038.96(3.9)+     \\
F16-D10 & 81.81(2)          & 81.88(2)          & 81.98(2.1)         & 82.46(2)          & 82.42(2.1)        & 82.37(2.7)        & 82.88(2.1)+       & 81.85(2.2)        & 82.29(2.5)        \\
F17-D10 & -16.92(0.0064)    & -16.90(0.012)+    & -16.92(0.0059)-    & -16.92(0.004)     & -16.91(0.0088)+   & -16.89(0.014)+    & -16.90(0.011)+    & -16.87(0.019)+    & -16.86(0.024)+    \\
F18-D10 & -16.85(0.022)     & -16.81(0.034)+    & -16.88(0.018)-     & -16.87(0.023)-    & -16.84(0.03)      & -16.78(0.058)+    & -16.82(0.043)+    & -16.73(0.042)+    & -16.65(0.061)+    \\
F19-D10 & -99.79(0.44)      & -99.71(0.57)      & -99.98(0.61)       & -99.96(0.56)      & -99.89(0.42)      & -99.72(0.58)      & -99.92(0.59)      & -99.73(0.67)      & -99.54(0.54)      \\
F20-D10 & -544.30(0.21)     & -544.25(0.17)     & -544.19(0.15)+     & -544.28(0.2)      & -544.29(0.15)     & -544.34(0.18)     & -544.29(0.12)     & -544.30(0.17)     & -544.23(0.22)     \\
F21-D10 & 42.05(1.1)        & 42.06(0.89)       & 42.26(0.94)        & 42.42(1.2)        & 42.69(1.7)        & 42.47(1.3)        & 42.48(1.1)        & 42.14(0.95)       & 41.90(0.7)        \\
F22-D10 & -998.15(0.4)      & -998.18(0.49)     & -998.09(0.36)      & -998.15(0.4)      & -998.13(0.32)     & -998.09(0.37)+    & -998.07(0.26)+    & -998.42(0.72)     & -998.02(0.11)+    \\
F23-D10 & 8.45(0.26)        & 8.50(0.36)        & 8.50(0.26)         & 8.50(0.24)        & 8.45(0.33)        & 8.53(0.34)        & 8.56(0.39)        & 8.45(0.39)        & 8.53(0.31)        \\
F24-D10 & 150.67(6)         & 150.68(4.8)       & 147.59(5.8)-       & 147.37(5.3)       & 148.99(5.7)       & 149.77(5.7)       & 149.32(6.4)       & 148.93(5.7)       & 151.04(6.1)       \\
\hline
F1-D20  & 79.59(0.17)       & 79.57(0.24)       & 79.63(0.23)        & 79.57(0.15)       & 79.59(0.2)        & 79.55(0.16)       & 79.57(0.17)       & 79.83(0.38)+      & 79.81(0.42)+      \\
F2-D20  & -70.60(1.4e+02)   & -89.61(1.3e+02)   & -85.18(95)         & -76.75(1.1e+02)   & -123.78(78)       & -119.19(82)       & -102.68(83)       & 58.38(2.2e+02)+   & 31.49(1.3e+02)+   \\
F3-D20  & -337.32(13)       & -329.33(14)+      & -331.08(15)        & -332.08(14)       & -334.18(14)       & -331.15(13)       & -337.10(14)       & -328.32(10)+      & -332.54(12)       \\
F4-D20  & -293.28(22)       & -293.61(22)       & -288.49(24)        & -290.59(33)       & -299.96(23)       & -288.69(26)       & -296.57(20)       & -282.26(27)       & -281.49(19)+      \\
F5-D20  & 14.28(11)         & 12.42(10)         & 12.26(10)          & 15.00(12)         & 10.95(9.5)        & 14.64(14)         & 15.28(12)         & 13.41(12)         & 42.81(15)+        \\
F6-D20  & 45.98(6.4)        & 43.77(4.7)        & 44.29(7.8)         & 43.71(3.7)        & 45.17(7.6)        & 43.66(3.7)        & 52.33(36)         & 47.84(5)          & 50.67(5.5)+       \\
F7-D20  & 95.45(2.5)        & 95.55(3.2)        & 96.90(3.8)         & 95.85(3.6)        & 95.24(1.9)        & 95.39(1.6)        & 96.33(3.6)        & 98.59(5.4)+       & 95.92(1.7)        \\
F8-D20  & 195.49(33)        & 184.84(27)        & 187.96(27)         & 182.45(21)        & 189.76(26)        & 189.15(26)        & 188.65(33)        & 220.88(59)+       & 208.40(40)        \\
F9-D20  & 144.20(3.8)       & 142.95(2.3)       & 142.82(2.6)        & 143.32(4.3)       & 144.42(3.7)       & 146.54(3.3)+      & 144.47(2.7)       & 144.56(9.9)       & 152.46(6.3)+      \\
F10-D20 & 190.42(3.3e+02)   & 153.54(1.8e+02)   & 217.85(3.1e+02)    & 188.35(3.6e+02)   & 80.89(2.7e+02)-   & 92.81(1.4e+02)    & 198.17(4.3e+02)   & 348.48(3.8e+02)+  & 786.24(7.9e+02)+  \\
F11-D20 & 77.58(1.6)        & 77.63(2.7)        & 77.96(1.9)         & 77.54(1.6)        & 77.34(1.4)        & 76.82(0.52)-      & 76.89(1)-         & 78.42(1.8)+       & 78.58(1.7)+       \\
F12-D20 & 8.2e+04(1.57e+05) & 20908.88(2.6e+04) & 43893.79(7.9e+04)  & 56096.68(8.3e+04) & 28508.42(4.9e+04) & 16612.98(1.2e+04) & 33022.39(8.8e+04) & 1.3e+05(1.39e+05) & 2e+05(2.95e+05)+  \\
F13-D20 & 66.03(30)         & 60.34(23)         & 84.32(53)          & 72.17(29)         & 64.29(27)         & 59.02(23)         & 58.23(21)         & 105.43(67)+       & 88.02(31)+        \\
F14-D20 & -52.29(0.055)     & -52.22(0.28)      & -52.20(0.25)       & -52.29(0.068)     & -52.26(0.13)      & -52.31(0.03)      & -52.22(0.22)      & -52.16(0.16)+     & -52.15(0.18)+     \\
F15-D20 & 1129.57(14)       & 1127.64(13)       & 1128.41(14)        & 1131.32(16)       & 1133.32(15)       & 1139.00(18)       & 1132.43(17)       & 1138.21(18)+      & 1134.00(14)       \\
F16-D20 & 93.46(2.7)        & 92.83(2.5)        & 93.66(2.4)         & 93.95(2.5)        & 93.10(2.8)        & 93.83(2.9)        & 93.48(2.3)        & 92.47(3.3)        & 94.07(2.2)        \\
F17-D20 & -16.30(0.39)      & -16.15(0.41)      & -16.23(0.38)       & -16.23(0.46)      & -16.31(0.38)      & -16.14(0.37)      & -16.25(0.36)      & -16.05(0.28)+     & -16.09(0.31)+     \\
F18-D20 & -14.58(0.98)      & -14.50(1.2)       & -14.94(1.4)        & -14.85(1.3)       & -15.06(0.82)      & -14.61(0.76)      & -14.77(0.91)      & -14.04(1.2)       & -13.79(1.5)+      \\
F19-D20 & -97.53(0.41)      & -97.25(0.5)+      & -97.73(0.43)       & -97.42(0.31)      & -97.57(0.44)      & -97.24(0.51)+     & -97.55(0.46)      & -97.40(0.48)      & -97.45(0.41)      \\
F20-D20 & -543.60(0.12)     & -543.59(0.14)     & -543.54(0.11)      & -543.63(0.15)     & -543.58(0.13)     & -543.56(0.11)     & -543.61(0.16)     & -543.59(0.11)     & -543.54(0.14)     \\
F21-D20 & 45.78(5.1)        & 46.12(5.6)        & 44.32(4.1)         & 45.63(5.4)        & 44.57(3.6)        & 46.08(5.5)        & 44.84(4.8)        & 44.94(4.5)        & 44.31(3.1)        \\
F22-D20 & -997.61(2.3)      & -997.03(4.2)      & -995.80(5.1)+      & -997.72(2.3)      & -996.54(4.6)      & -995.12(6.2)      & -997.23(3.2)      & -996.84(3.3)+     & -994.82(8)+       \\
F23-D20 & 9.39(0.48)        & 9.51(0.41)        & 9.50(0.41)         & 9.44(0.43)        & 9.52(0.35)        & 9.39(0.52)        & 9.37(0.39)        & 9.49(0.38)        & 9.53(0.39)        \\
F24-D20 & 246.17(9.4)       & 243.25(9.2)       & 238.53(9.9)-       & 242.01(9.3)       & 243.87(11)        & 247.67(9.5)       & 246.22(12)        & 240.23(13)        & 250.83(10)\\
\bottomrule
    \end{tabular}
    }
    \end{adjustbox}
    \label{tab:bbob2023}
\end{table*}

\begin{table*}[h]
    \centering
    \caption{Output solution's average objective value and standard deviation from 30 runs on the Solomon CVRPTW C-type and RC-type instances. "+" or "-" indicate the objective values are significantly larger (worse) or smaller (better) than objective values obtained by HF, respectively.}
    \begin{adjustbox}{width=\textwidth,center}
     \setlength\tabcolsep{1pt}{
    \begin{tabular}{c|ccccccccc}
\toprule
  & HF (Ours)                  & HF-NU        & HF-NA(0.5)         & HF-NA(0.3)         & HF-NA(0.1)       & DQN-GSF     & DQN-GSF-U             & SL                 & Random                \\
    \midrule
C101  & 18430.89(1050.54) & 18587.71(911.50)  & 18565.14(985.40)+ & 18493.77(1020.96)+ & 18602.84(954.35)+  & 18648.22(926.95)  & 18457.22(1002.19)  & 18717.17(965.08)+  & 18734.49(999.33)+  \\
C102  & 17182.90(729.61)  & 17749.83(715.35)+ & 17217.70(682.73)  & 17255.91(708.98)+  & 17569.18(741.26)+  & 18128.85(743.59)+ & 17309.77(740.65)+  & 17288.55(696.01)+  & 17282.94(694.62)+  \\
C103  & 15630.23(556.20)  & 15682.12(598.47)  & 15673.26(559.67)+ & 15632.75(587.30)   & 15657.21(561.84)   & 15841.84(599.20)  & 15705.17(630.56)   & 15727.62(571.57)+  & 15742.39(535.40)+  \\
C104  & 13033.17(502.73)  & 12984.19(495.19)  & 13090.72(466.35)+ & 13070.74(487.60)+  & 13078.32(487.47)   & 13032.28(484.48)  & 13109.38(506.63)+  & 13160.28(483.10)+  & 13157.28(480.44)+  \\
C105  & 15871.08(740.74)  & 16007.32(810.09)  & 15977.08(782.28)+ & 15942.80(782.41)+  & 16086.26(804.68)+  & 16121.94(801.81)  & 15905.75(826.58)   & 16151.83(703.68)+  & 16155.52(702.83)+  \\
C106  & 16081.24(1020.99) & 16370.22(1012.68) & 16145.53(983.57)  & 16149.34(1031.39)+ & 16361.90(1022.58)+ & 16536.35(1012.62) & 16186.72(1027.09)+ & 16276.71(1051.36)+ & 16313.00(1016.44)+ \\
C107  & 13284.36(557.18)  & 13621.77(578.14)  & 13340.45(605.16)+ & 13356.60(585.72)+  & 13545.59(591.69)+  & 13783.28(618.33)+ & 13421.27(626.56)+  & 13454.36(591.65)+  & 13434.16(570.95)+  \\
C108  & 13541.03(537.49)  & 13762.48(489.71)  & 13818.68(635.42)+ & 13627.24(491.14)+  & 13977.25(541.54)+  & 14107.01(475.10)+ & 13610.84(538.24)+  & 13492.01(506.80)   & 13515.42(501.83)   \\
C109  & 12978.22(550.48)  & 12830.36(541.20)  & 13475.59(457.82)+ & 13090.02(518.46)+  & 13579.03(434.91)+  & 13509.66(414.83)+ & 13059.65(502.62)   & 11976.51(462.46)-  & 11980.05(439.09)-  \\
C201  & 5629.56(637.59)   & 5852.99(587.99)   & 6614.18(678.26)+  & 5637.19(616.92)    & 6607.03(677.94)+   & 6072.83(629.36)+  & 5666.10(637.40)    & 5725.68(628.80)+   & 5765.90(612.07)+   \\
C202  & 5941.01(467.46)   & 6240.92(419.93)+  & 7700.00(495.08)+  & 5962.10(472.11)    & 7587.60(501.44)+   & 6586.41(357.11)+  & 6026.71(475.04)+   & 6072.74(478.93)+   & 6110.24(473.02)+   \\
C203  & 5572.13(468.11)   & 6853.78(624.18)+  & 7842.99(418.28)+  & 5624.40(434.01)+   & 7938.84(398.03)+   & 7657.80(462.09)+  & 5708.67(496.95)+   & 5623.76(499.28)+   & 5643.24(469.14)+   \\
C204  & 4405.54(199.23)   & 4429.29(204.58)   & 6791.08(220.80)+  & 4426.75(189.82)    & 6853.83(231.10)+   & 4500.73(163.61)+  & 4491.20(215.33)+   & 4492.39(163.90)+   & 4524.70(172.88)+   \\
C205  & 4739.92(492.75)   & 4989.14(492.71)+  & 4781.76(510.22)   & 4749.03(494.33)    & 4891.37(507.44)+   & 5136.70(459.49)+  & 4758.75(487.14)    & 4834.26(473.27)+   & 4867.62(490.57)+   \\
C206  & 4440.04(545.79)   & 4614.61(488.90)   & 4528.54(496.82)+  & 4452.73(500.29)    & 4585.76(511.22)+   & 4746.29(477.96)   & 4485.16(519.54)    & 4727.10(488.12)+   & 4740.32(495.11)+   \\
C207  & 4600.25(442.65)   & 4819.40(445.64)   & 4581.84(437.39)   & 4609.40(458.87)    & 4735.78(446.41)+   & 4961.91(454.44)+  & 4644.32(444.51)+   & 4590.03(449.10)    & 4602.49(444.18)    \\
C208  & 4132.78(477.69)   & 4325.96(468.89)   & 4125.61(475.02)   & 4146.85(457.71)    & 4237.89(435.02)+   & 4458.02(432.88)+  & 4210.04(508.48)+   & 4194.17(447.81)+   & 4214.44(483.24)+   \\
\hline
RC101 & 12086.98(394.53)  & 12240.64(360.76)  & 12181.11(426.59)+ & 12123.23(392.62)   & 12281.19(452.99)+  & 12293.63(384.63)  & 12234.90(363.27)+  & 12256.40(374.36)+  & 12276.22(413.11)+  \\
RC102 & 11924.84(296.05)  & 12248.98(268.99)+ & 11992.29(311.91)+ & 12002.84(272.04)+  & 12198.73(241.21)+  & 12415.28(224.30)+ & 12093.50(252.92)+  & 12038.79(297.83)+  & 12073.56(276.41)+  \\
RC103 & 11242.55(285.10)  & 11599.08(273.02)+ & 11257.49(304.70)  & 11288.79(290.04)+  & 11528.63(315.68)+  & 11808.69(269.54)+ & 11400.07(289.31)+  & 11317.54(307.39)+  & 11324.57(288.09)+  \\
RC104 & 11012.86(217.07)  & 11400.78(207.07)+ & 11007.08(218.30)  & 11046.35(235.81)   & 11361.59(226.55)+  & 11633.92(197.70)+ & 11203.52(242.67)+  & 11018.33(225.76)   & 11036.38(218.69)   \\
RC105 & 13018.06(459.11)  & 13247.33(446.56)  & 13100.62(476.97)+ & 13091.80(441.78)+  & 13224.34(442.18)+  & 13377.20(445.41)+ & 13160.96(472.38)+  & 13137.06(472.73)+  & 13160.38(453.92)+  \\
RC106 & 11176.67(79.23)   & 11419.08(106.65)+ & 11228.72(100.70)+ & 11248.71(111.63)+  & 11337.01(111.00)+  & 11539.12(101.29)+ & 11304.71(148.12)+  & 11299.12(114.89)+  & 11315.50(94.77)+   \\
RC107 & 11040.23(79.40)   & 11329.38(110.29)+ & 11096.17(91.64)+  & 11068.32(66.73)    & 11254.38(116.99)+  & 11466.71(81.83)+  & 11119.32(100.76)+  & 11112.13(79.54)+   & 11121.66(68.42)+   \\
RC108 & 11074.58(166.89)  & 11221.60(90.55)+  & 11387.36(248.55)+ & 11046.77(146.85)   & 11429.58(184.49)+  & 11383.08(74.75)+  & 11094.58(114.68)   & 10930.44(76.04)-   & 10917.82(69.31)-   \\
RC201 & 5091.57(443.00)   & 4529.68(186.04)-  & 4111.73(68.95)-   & 5218.09(634.14)    & 4314.08(185.42)-   & 4819.43(174.86)-  & 4072.46(104.31)-   & 4241.01(80.12)-    & 4186.42(74.03)-    \\
RC202 & 6018.94(181.46)   & 4283.44(106.50)-  & 3918.43(69.13)-   & 6114.22(134.14)+   & 4177.22(133.27)-   & 4433.68(106.64)-  & 3924.69(114.96)-   & 4005.16(93.76)-    & 3979.46(82.66)-    \\
RC203 & 5801.11(208.93)   & 3867.76(84.54)-   & 3671.09(63.96)-   & 5873.07(155.98)    & 3870.98(92.04)-    & 4005.09(61.40)-   & 3615.96(83.99)-    & 3712.11(71.72)-    & 3747.90(89.72)-    \\
RC204 & 5890.94(244.27)   & 3560.42(78.35)-   & 3429.58(58.97)-   & 5910.37(195.09)    & 3569.69(89.09)-    & 3631.25(66.95)-   & 3402.50(64.60)-    & 3483.75(85.73)-    & 3493.81(83.17)-    \\
RC205 & 6142.27(163.77)   & 4329.59(123.48)-  & 4042.26(69.97)-   & 6155.42(142.79)    & 4210.86(185.93)-   & 4607.42(101.54)-  & 3985.97(90.37)-    & 4109.44(76.09)-    & 4134.00(78.73)-    \\
RC206 & 3634.38(71.36)    & 3974.15(89.07)+   & 3720.07(60.61)+   & 3703.43(92.51)+    & 3882.16(147.42)+   & 4148.28(79.61)+   & 3671.42(98.76)     & 3784.67(77.65)+    & 3813.85(93.27)+    \\
RC207 & 3602.63(74.42)    & 3827.87(88.78)+   & 3742.37(116.71)+  & 3628.09(68.11)     & 3965.74(129.10)+   & 3945.16(61.03)+   & 3615.91(73.53)     & 3755.08(90.67)+    & 3760.82(95.79)+    \\
RC208 & 3348.33(55.46)    & 3536.89(48.43)+   & 3933.19(248.21)+  & 3376.08(64.77)     & 4077.95(223.34)+   & 3624.84(47.84)+   & 3477.45(97.68)+    & 3474.96(68.87)+    & 3471.27(76.59)+\\
\bottomrule
    \end{tabular}
    }
    \end{adjustbox}
    \label{tab:C_RC}
\end{table*}

\end{document}